\documentclass[10pt,twocolumn,letterpaper]{article}

\usepackage[pagenumbers]{cvpr} 

\usepackage{amsmath,amsfonts,bm}

\def\eqref#1{equation~\ref{#1}}

\def\1{\bm{1}}

\DeclareMathAlphabet{\mathsfit}{\encodingdefault}{\sfdefault}{m}{sl}
\SetMathAlphabet{\mathsfit}{bold}{\encodingdefault}{\sfdefault}{bx}{n}

\DeclareMathOperator*{\argmin}{arg\,min}

\DeclareMathOperator{\Tr}{Tr}

\usepackage{graphicx}
\usepackage{amsmath,amssymb, bm, bbm}
\usepackage{amssymb}
\usepackage{booktabs}
\usepackage{multirow}
\usepackage{textcomp}

\newtheorem{theorem}{Theorem}

\newtheorem{lemma}{Lemma}

\usepackage[pagebackref,breaklinks,colorlinks]{hyperref}

\usepackage[capitalize]{cleveref}
\crefname{section}{Sec.}{Secs.}
\Crefname{section}{Section}{Sections}
\Crefname{table}{Table}{Tables}
\crefname{table}{Tab.}{Tabs.}

\def\cvprPaperID{12063} 
\def\confName{CVPR}
\def\confYear{2023}

\newcommand\blfootnote[1]{%
  \begingroup
  \renewcommand\thefootnote{}\footnote{#1}%
  \addtocounter{footnote}{-1}%
  \endgroup
}

\begin{document}

\title{CUDA: Convolution-based Unlearnable Datasets}

\author{Vinu Sankar Sadasivan\\
University of Maryland\\
{\tt\small vinu@umd.edu}
\and
Mahdi Soltanolkotabi\\
University of Southern California\\
{\tt\small soltanol@usc.edu}
\and
Soheil Feizi\\
University of Maryland\\
{\tt\small sfeizi@cs.umd.edu}\\
}

\maketitle

\begin{abstract}
   Large-scale training of modern deep learning models heavily relies on publicly available data on the web. This potentially unauthorized usage of online data leads to concerns regarding data privacy. Recent works aim to make unlearnable data for deep learning models by adding small, specially designed noises to tackle this issue. However, these methods are vulnerable to adversarial training (AT) and/or are computationally heavy. In this work, we propose a novel, model-free, Convolution-based Unlearnable DAtaset (CUDA) generation technique. CUDA is generated using controlled class-wise convolutions with filters that are randomly generated via a private key. CUDA encourages the network to learn the relation between filters and labels rather than informative features for classifying the clean data. We develop some theoretical analysis demonstrating that CUDA can successfully poison Gaussian mixture data by reducing the clean data performance of the optimal Bayes classifier. We also empirically demonstrate the effectiveness of CUDA with various datasets (CIFAR-10, CIFAR-100, ImageNet-100, and Tiny-ImageNet), and architectures (ResNet-18, VGG-16, Wide ResNet-34-10, DenseNet-121, DeIT, EfficientNetV2-S, and MobileNetV2). Our experiments show that CUDA is robust to various data augmentations and training approaches such as smoothing, AT with different budgets, transfer learning, and fine-tuning. For instance, training a ResNet-18 on ImageNet-100 CUDA achieves only 8.96$\%$, 40.08$\%$, and 20.58$\%$ clean test accuracies with empirical risk minimization (ERM), $L_{\infty}$ AT, and $L_{2}$ AT, respectively. Here, ERM on the clean training data achieves a clean test accuracy of 80.66$\%$. CUDA exhibits unlearnability effect with ERM even when only a fraction of the training dataset is perturbed. Furthermore, we also show that CUDA is robust to adaptive defenses designed specifically to break it.
\end{abstract}


\section{Introduction} \label{sec:introduction}

Modern deep learning training frameworks heavily depend on large-scale datasets for achieving high accuracy. This encourages deep learning practitioners to scrape data from the web for data collection \cite{largedata1, largedata2, largedata3, largedata4}. Since a lot of the data is publicly available online, sometime this scrapping of data is unauthorized. For instance, a recent article \cite{hillarticle} discloses that a private company trained a commercial face recognition system using over three billion facial images collected from the internet without any user consent. Although such massive data can significantly boost the performance of deep learning models, it raises serious concerns about data privacy and security.

To prevent the unauthorized usage of personal data, a series of recent papers \cite{em, tap, ntga} propose to poison data with additive noise. The idea is to make datasets {\it unlearnable} for deep learning models by ensuring that they learn the correspondence between noises and labels. Thereby, they do not learn much useful information about the clean data, significantly degrading their clean test accuracy. However, in recent works \cite{em, betterthansorry, rem}, these unlearnability methods are shown to be vulnerable to adversarial training (AT) frameworks \cite{adversarialtraining}. Motivated by this problem, Fu \textit{et al.} \cite{rem} developed Robust Error-Minimization (REM) noises to make unlearnable data that is protected from AT. While the authors show the effectiveness of REM in multiple scenarios, we demonstrate that these methods are still not robust against different data augmentations or training settings (see Section \ref{sec:limitations}). Furthermore, current unlearnability frameworks \cite{rem, em, tap, ntga} are model-dependent and require expensive optimization steps on deep learning models to obtain the additive noises. They also need to train the deep learning models from scratch to obtain noises for each new data set.\blfootnote{Code available here: \url{https://github.com/vinusankars/Convolution-based-Unlearnability}}

 \begin{figure*}[t]
 \centering
     \includegraphics[width=0.85\textwidth]{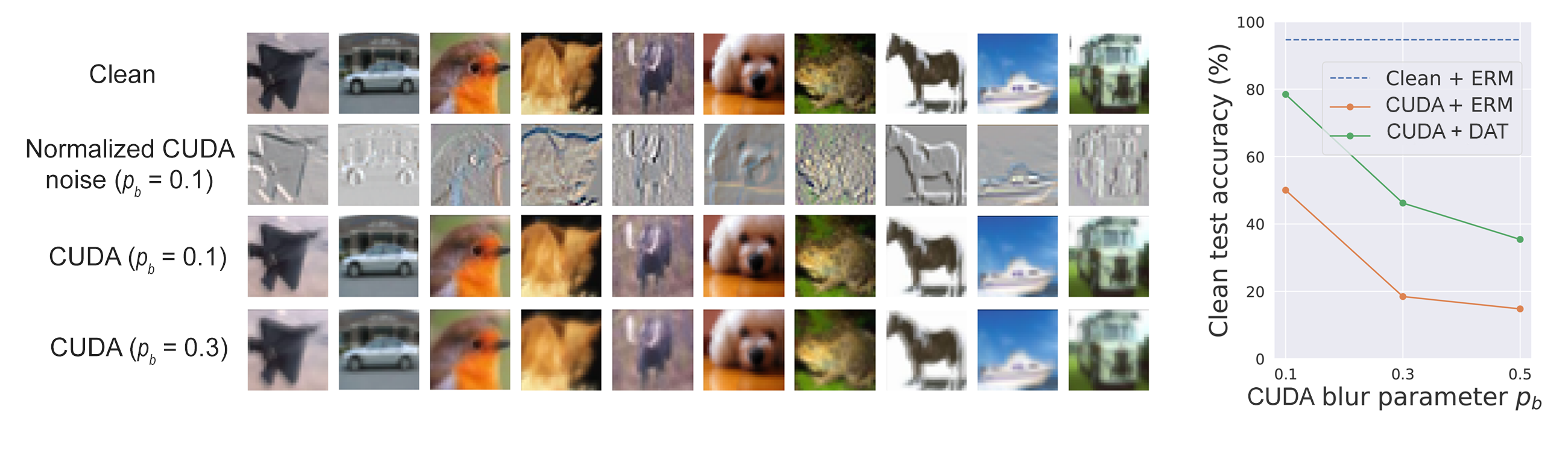}
     \caption{CIFAR-10 CUDA images from each of the 10 classes convolved using convolutional filters generated with blur parameter $p_b$. As seen in the plot, a higher $p_b$ value results in better unlearnability (lower clean test accuracy), but increased blurring. CIFAR-10 CUDA does not break with ERM for all the three $p_b$ values. In Section \ref{sec:effect_fun}, we propose Deconvolution-based Adversarial Training (DAT) that we specifically design to break CUDA. DAT adversarially learns class-wise filters to deconvolve CUDA images. However, CIFAR-10 CUDA with $p_b \in \{0.3, 0.5\}$ does not break with DAT. Therefore, we select $p_b=0.3$ (lesser blurring) for CIFAR-10 CUDA as discussed in Section \ref{sec:expsetup}. More details on DAT in supplementary material.}
     \label{fig:funcifar}
 \end{figure*}

\begin{figure*}[t]
 \centering
     \includegraphics[width=0.75\textwidth]{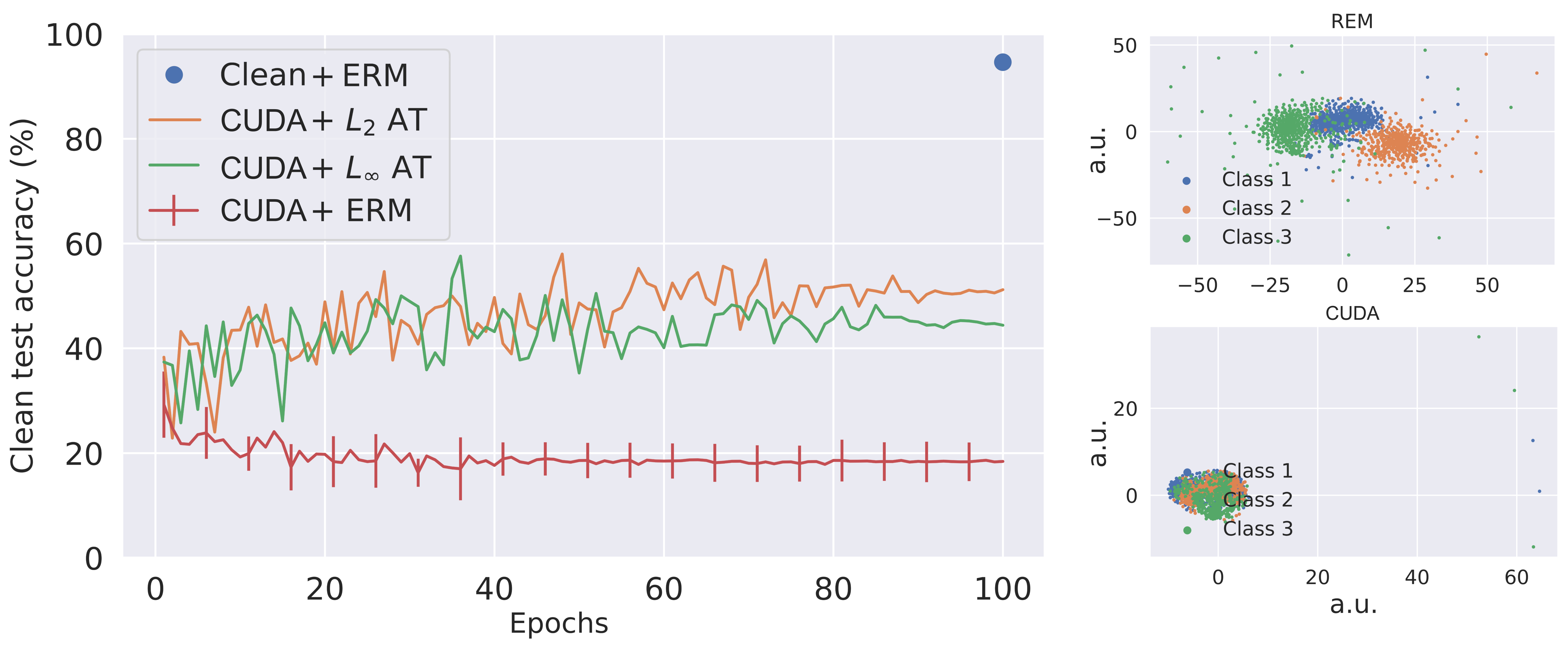}
     \caption{In the left column, clean test accuracies of ResNet-18 on clean CIFAR-10 and CIFAR-10 CUDA are shown. The lower the test accuracy, the higher the effectiveness of unlearnability is. We show that CUDA is robust to various training settings such as ERM, $L_{2}$ AT, and $L_{\infty}$ AT. The error bars represent the standard deviation of test accuracy over 5 independent trials. CUDA is robust to the randomness in its data generation process. In the right column, t-SNEs of the noises generated by REM (top) and CUDA (bottom) for the first 3 classes of CIFAR-10 are plotted.}
     \label{fig:title}
 \end{figure*}

In this paper, we propose a novel Convolution-based Unlearnable DAtaset (CUDA) generation technique. We address limitations of existing unlearnable data generation techniques in Section \ref{sec:limitations} and motivate our CUDA technique in Section \ref{sec:fun}. For generating CUDA, an attacker \textit{randomly generates} different convolutional filters for each class in the dataset using {\it a private key} or seed value. These filters are used to perform {\it controlled class-wise convolutions} on the clean training dataset to obtain CUDA. As we describe in Sections \ref{sec:fun} and \ref{sec:effect_fun}, CUDA generation performs controlled convolutions using a blur parameter $p_b$ to ensure that the semantics of the dataset are preserved (see Figure \ref{fig:funcifar}). CUDA generation with a lower blurring parameter $p_b$ adds less perceptible noises to clean samples.  A network trained by a defender on CUDA is encouraged to learn the \textit{shortcut} relation between class-wise convolutional filters and labels rather than useful features for classifying the clean data. Since the seed value for generating the filters are private, its not possible for the defender to obtain clean data from CUDA alone. Additionally, CUDA exhibits unlearnability effect with ERM even when only a fraction of the training dataset is perturbed (see Section \ref{sec:experiments}). While the existing unlearnability  works use {\it additive} noises, CUDA generation technique enjoys the advantage of introducing {\it multiplicative} noises in the Fourier domain due to the convolution theorem (since convolution of signals is the same as element-wise multiplication in the Fourier domain). This lets CUDA generation add higher amounts of noise in the image space, specifically along the edges in images, and makes it resilient to AT with small additive noise budgets. In Figure \ref{fig:title}, with the help of t-SNE plots \cite{tsne}, we also find that the noises added by CUDA generation is {\it not} linearly separable while they are linearly separable for the existing works on unlearnability \cite{indiscriminateshortcut}.

In Section \ref{sec:theory}, we theoretically show that CUDA generation can successfully poison Gaussian mixture data by degrading the clean data accuracy of the optimal Bayes classifier. We state our result informally below while the formal version is presented in Theorem \ref{theorem:funbound}.

\begin{theorem}[Informal]
\label{theorem:informal}
Let $\mathcal{D}$ denote a Gaussian mixture data with two modes, $P_{\mathcal{D}}$ denote the optimal Bayes classifier trained on $\mathcal{D}$, and $\tau_{\mathcal{D}}(P_{\mathcal{D}})$ denote the accuracy of the classifier $P_{\mathcal{D}}$ on $\mathcal{D}$. Then, under some assumptions, for every clean data $\mathcal{D}$, there is a CUDA $\mathcal{\tilde{D}}$ such that $\tau_{\mathcal{D}}(P_{\mathcal{\tilde{D}}}) < \tau_{\mathcal{D}}(P_{\mathcal{D}})$.
\end{theorem}

Furthermore, our empirical experiments in Section \ref{sec:experiments} demonstrate the effectiveness of CUDA under various training scenarios such as ERM with various augmentations and regularizations, AT with different budgets, randomized smoothing \cite{smoothingcohen, smoothingpoison, smoothingcertificate}, transfer learning, and fine-tuning. For instance, training a ResNet-18 on CIFAR-10 CUDA achieves only 18.48$\%$, 44.4$\%$, and 51.14$\%$ clean test accuracies with ERM, $L_{\infty}$ AT, and $L_{2}$ AT, respectively (see Figure \ref{fig:title}). Here, ERM on the clean training data achieves a clean test accuracy of 94.66$\%$. In addition, we also design adaptive defenses to investigate if CUDA breaks with random or adversarial defense mechanisms. We find that CUDA is robust to the adaptive defenses that we specifically design to break it.

\section{Related Works} \label{sec:relatedworks}
Our CUDA generation technique is intimately related with adversarial and poisoning attacks. We first discuss some of this literature and then explain their relation with CUDA generation.

\textbf{Adversarial attacks.} Adversarial examples are specially designed examples that can fool deep learning models at test time \cite{adveg1, adveg2, adveg3, adveg4, funcattacks}. The adversary crafts these examples by adding error-maximizing noises to the clean data. Even slightly perturbed data can serve as adversarial examples. AT is a training framework proposed to make deep learning models robust to adversarial examples \cite{adversarialtraining, at2, at4, par, at3}. AT is a min-max optimization problem where the model is trained to minimize loss on adversarial examples that have the maximum loss.

\textbf{Poisoning attacks.} In data poisoning, an attacker aims to hurt the deep learning model's performance by perturbing the training data \cite{poisoning1, poisoning2, poisoning3, dpa, poisoning4, dfa}. The backdoor attack is a special type of poisoning attack where a trigger pattern is injected into clean data at training time \cite{backdoor1, backdoor2, backdoor3, backdoor4}. The model trained on this data would misclassify an image with a trigger pattern at test time. Gu \textit{et al.} \cite{backdoor6} and Li \textit{et al.} \cite{backdoor7} use perceptible amounts of noises similar to CUDA for data poisoning. However, backdoor attacks do not affect the performance of the model on clean data \cite{backdoor1, poisoning3, backdoor5}.

Recent literature utilize data poisoning to protect data from being used for model training without authorization. Yuan and Wu \cite{ntga} use neural tangent kernels \cite{ntk} to generate clean label attacks that can hurt the generalization of deep learning models. Huang \textit{et al.} \cite{em} show that error-minimizing noise addition can serve as a poisoning technique. Fowl \textit{et al.} \cite{tap} show that error-maximizing noises as well can make strong poison attacks. However, all these poisoning techniques do not offer data protection with AT \cite{indiscriminateshortcut, betterthansorry}. Fu \textit{et al.} \cite{rem} proposes a min-min-max optimization technique to generate poisoned data that offers better unlearnability effects with AT.
\section{Convolution-based Ulearnable DAtaset (CUDA)}
\label{sec:fun}

In this section, first we give some preliminaries about unlearnability. Then we discuss the limitations of the existing unlearnability methods. Finally, we propose our CUDA generation technique.

\subsection{Preliminaries}
\label{sec:preliminaries}

Let $\{(\bm{x}_i, y_i)\}_{i=1}^n \sim \mathcal{D}^n$ be the clean training dataset where  $\mathcal{D}$ is the clean data distribution, $\bm{x}_i \in \mathcal{X} \subset \mathbb{R}^d$ are the samples, and $y_i \in \mathcal{Y}$ are the corresponding labels. Suppose a network is given as $f_{\theta}: \mathcal{X} \to \mathcal{Y}$ where $\theta \in \Theta$ is the network parameter and $\ell: \mathcal{Y} \times \mathcal{Y} \to \mathbb{R}$ is the loss function. ERM trains a network using a minimization problem of the form: $\min_{\theta} \frac{1}{n} \sum_{i=1}^n \ell(f_{\theta}(\bm{x}_i), y_i)$.
Standard $L_p$-based AT (for $p \in \mathbb{R}^+$)  solves the following min-max problem:     $\min_{\theta} \frac{1}{n} \sum_{i=1}^n \max_{\|\bm{\delta}_i\|_p \leq \rho_a} \ell(f_{\theta}(\bm{x}_i + \bm{\delta}_i), y_i)$
where $\rho_a$ is the adversarial perturbation radius.

We will use $\tau_{\mathcal{D}}(\theta)$, with $\theta$ the clean model parameter, to denote the clean test accuracy of a model trained on the clean training dataset (i.e., clean model accuracy). In unlearnable dataset generation, an attacker uses an algorithm $\mathcal{A}: \mathcal{X} \to \mathcal{X}$ to generate an unlearnable dataset $\{(\tilde{\bm{x}}_i = \mathcal{A}(\bm{x}_i), y_i)\}_{i=1}^n \sim \tilde{\mathcal{D}}^n$ from the clean training data. Here, the attacker assumes access to the full clean training datatset. Moreover, the attacker cannot modify the unlearnable dataset once it is released publicly.  A defender trains using the unlearnable dataset to obtain a network $f_{\tilde{\theta}}$. The objective of the attacker is to design an unlearnable dataset such that the defender's model trained on the unlearnable data achieves a clean test accuracy (i.e., unlearnable model accuracy) worse than the clean model accuracy i.e.~$\tau_{{\mathcal{D}}}(\tilde{\theta}) \ll \tau_{{\mathcal{D}}}({\theta})$.

\subsection{Limitations of existing works}
\label{sec:limitations}

Fu \textit{et al.} \cite{rem} show that the previous unlearnability methods including Error-Minimization (EM) \cite{em}, Targeted Adversarial Poisoning (TAP) \cite{tap}, and Neural Tangent Generalization Attack (NTGA) \cite{ntga} are vulnerable to AT. Hence, they propose a Robust Error-Minimization (REM) \cite{rem} method that exploits a min-min-max optimization procedure to generate unlearnable noises. REM first trains a noise generator $f'_{\theta}$ on data points $\{(\bm{x}_i, y_i)\}_{i=1}^n$ over a loss function $\ell$ as follows
\begin{align}
    \min_{\theta} ~&\frac{1}{n} \sum_{i=1}^n \min_{\|\bm{\delta}_i^u\| \leq \rho_u} \mathbb{E}_{t \sim T} \nonumber\\ &\max_{\|\bm{\delta}_i^a\| \leq \rho_a} \ell(f'_{\theta}(t(\bm{x}_i+ \bm{\delta}_i^u) + \bm{\delta}_i^a), y).
\end{align}
Here, $T$ is a distribution over a set of transformations $\{t: \mathcal{X} \to \mathcal{X}\}$, $\rho_u$ is the defensive perturbation radius, and $\rho_a$ controls the protection level of REM against AT. After training the noise generator,  an unlearnable example $(\bm{\tilde{x}}, y)$ is generated via
\begin{align}\label{opt:REM}
    \bm{\tilde{x}} = \bm{x} + \argmin_{\|\bm{\delta}^u\| \leq \rho_u} ~&\mathbb{E}_{t \sim T} \nonumber \\ \max_{\|\bm{\delta}^a\| \leq \rho_a} &\ell(f'_{\theta}(t(\bm{x}+ \bm{\delta}^u) + \bm{\delta}^a), y)
\end{align}

\begin{table}[t]
  \caption{Time taken for generating unlearnable noises for various datasets using different unlearnability techniques.}
  \label{tab:time}
  \footnotesize
  \centering
  \begin{tabular}{c l l l l l}
    \toprule
    Dataset & EM & TAP & NTGA & REM & \textbf{CUDA} \\
    \midrule
    CIFAR-10 & 0.4 hr & 0.5 hr & 5.2 hrs & 22.6 hrs & \textbf{10.8 s}\\
    CIFAR-100 & 0.4 hr & 0.5 hr & 5.2 hrs & 22.6 hrs & \textbf{15.5 s}\\
    ImageNet-100 & 3.9 hrs & 5.2 hrs & 14.6 hrs & 51.2 hrs & \textbf{0.15 hr}\\

    \bottomrule[0.25ex]
  \end{tabular}
\end{table}

First, note that REM is computationally expensive since it needs to generate unlearnability noises through solving optimization \eqref{opt:REM}. Moreover, the existing techniques are model-dependent and they require gradient-based training with a neural network to generate unlearnable data. They also require neural network training from scratch for every dataset that is to be made unlearnable. Table \ref{tab:time} shows the amount of time required to generate various unlearnable datasets using NVIDIA\textsuperscript{\textregistered} Tesla V100 GPU and 10 CPU cores. CUDA generation is significantly faster than the existing methods since it uses a model-free approach (no training required). Furthermore, REM is sensitive to hyperparameters and norm-budgets of AT since they generate noises with fixed $L_{\infty}$ norm budgets. For instance, a ResNet-18 trained on clean CIFAR-10 dataset achieves a clean test data accuracy of 94.66$\%$. $L_{\infty}$ AT with perturbation radius $\rho_a = 4/255$ on REM CIFAR-10 data (generated using $\rho_u = 8/255$ and $\rho_a = 4/255$) achieves only a clean test accuracy of 48.16$\%$. However, $L_{\infty}$ AT with perturbation radius $\rho_a = 8/255$ and $L_{2}$ AT with perturbation radius $\rho_a = 0.75$ can achieve a clean test accuracy of 78.71$\%$ and 79.65$\%$, respectively, on the same REM data. We also find that ERM with a ResNet-18 on grayscaled REM CIFAR-10 images can achieve a high test accuracy of 70.76$\%$ on the grayscaled CIFAR-10 test data. This shows that REM relies upon the color space for poisoning clean data. Fu \textit{et al.} \cite{rem} also show that REM noise generated using ResNet-18 is not transferable to DenseNet-121.

\subsection{Our method: CUDA}
\label{sec:fun}

The major limitations of the previous works are that they are vulnerable to AT, and computationally expensive. We think the major reason for the former limitation is the usage of small additive noises for unlearnability. AT is designed to train in the presence of such additive noises. Increasing the budget of the amount of additive noises for unlearnability might destroy the semantics of the images while perturbing them. The latter limitation arises from the fact that these methods are model-dependent and they require multilevel optimizations. Hence, we are motivated to design a compute-efficient unlearnability method that is robust to AT. CUDA technique can perform convolutions to add larger amounts of noises to clean images without destroying its semantics. This can help CUDA to be robust against AT. CUDA uses randomly generated convolution filters for blurring images from each class. This makes a model trained on CUDA to learn  \textit{shortcut} relations between filters and labels. We empirically support these claims in Section \ref{sec:effect_fun}. Additionally, randomly generating the filters makes our technique model-free. Since the keys for generating the filters are private, it is not possible to reverse the blurring effect in CUDA without having access to the corresponding clean images. Hence, we assume that the data publisher deletes the clean images after perturbing them. Moreover, CUDA technique is a novel class of non-additive noise based poisoning attack that needs to be studied.

CUDA uses convolutional filters $s_i \in \mathbb{R}^{k \times k}$ for each class $i \in [1,K]$. A random parameter, out of the $k^2$ parameters, in each of the filters is set to have a value of 1. The rest of the filter parameters are randomly initialized from a uniform distribution $\mathcal{U}(0, p_b)$ using a private seed where $p_b$ is the blur parameter.  Blur parameter controls the level of blurring that occurs when an image $\bm{x} \in [0,1]^{d_1 \times d_2 \times d_3}$ is convolved with a filter $s_i$. Here, $d_1, d_2$, and $d_3$ are the height, width, and number of channels of the image, respectively. For example, a CIFAR-10 image has a dimension of $32\times 32 \times 3$. The higher $p_b$ is, the higher the blurring effect is.  Let $\hat{\bm{x}} = \bm{x}*s_i$ where $\bm{x}$ belongs to class $i$. The CUDA data point for $\bm{x}$ is given by $\bm{\tilde{x}} = \hat{\bm{x}}/\textrm{MAX}(\hat{\bm{x}})$. Rescaling is performed to make sure that the CUDA image pixels lie between 0 and 1. We find that the unlearnability effect gets stronger with larger $p_b$ and $k$ values (see supplementary material for details).
\section{Theory for CUDA} \label{sec:theory}

\begin{figure*} 
    \centering
  \subfloat[$\|\bm{\mu}\|_2 = 1.0$\label{fig:theory1}]{%
       \includegraphics[width=0.2\linewidth]{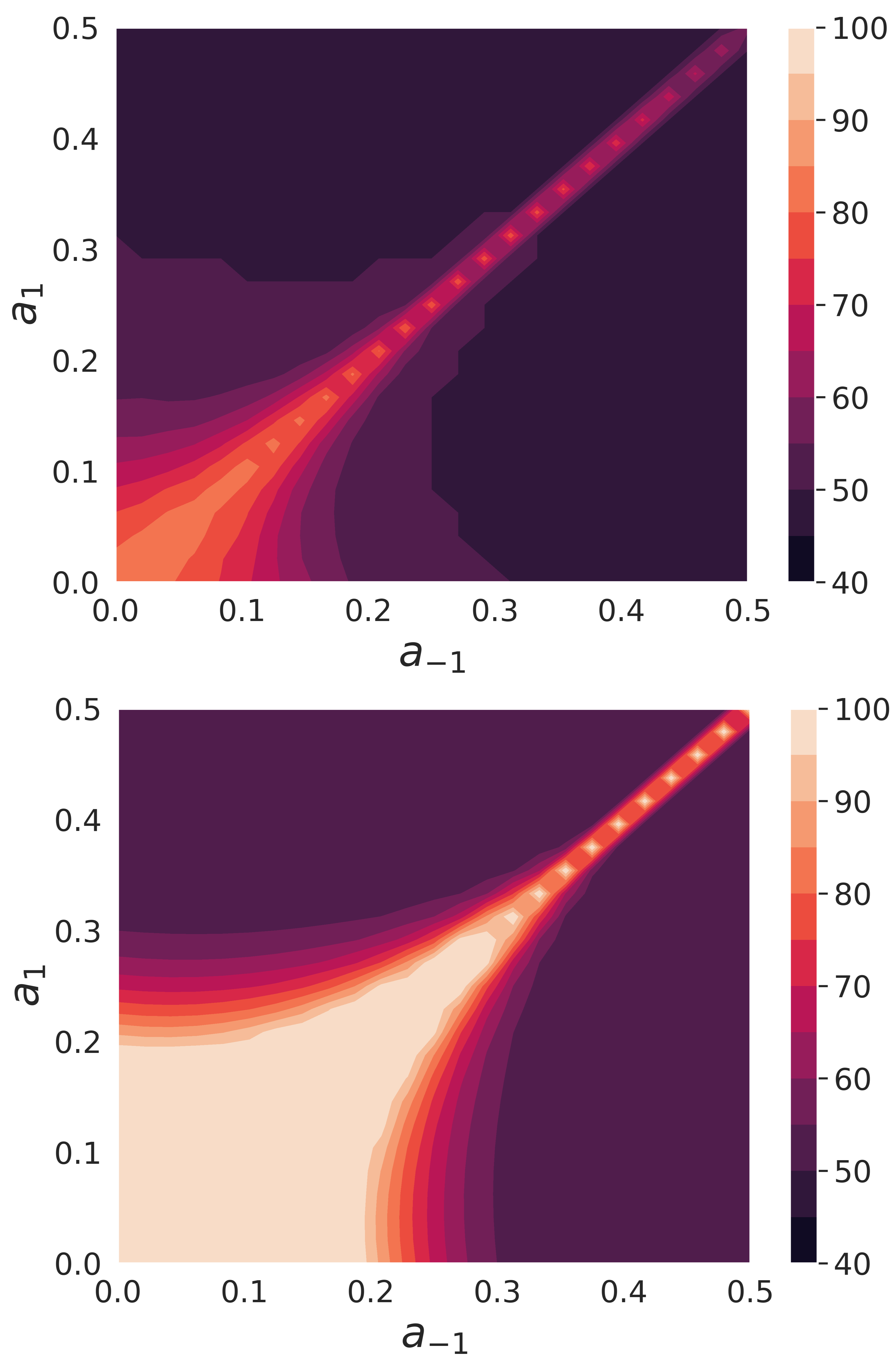}}
    \hspace{0.25cm}
  \subfloat[$\|\bm{\mu}\|_2 = 2.5$\label{fig:theory2}]{%
        \includegraphics[width=0.2\linewidth]{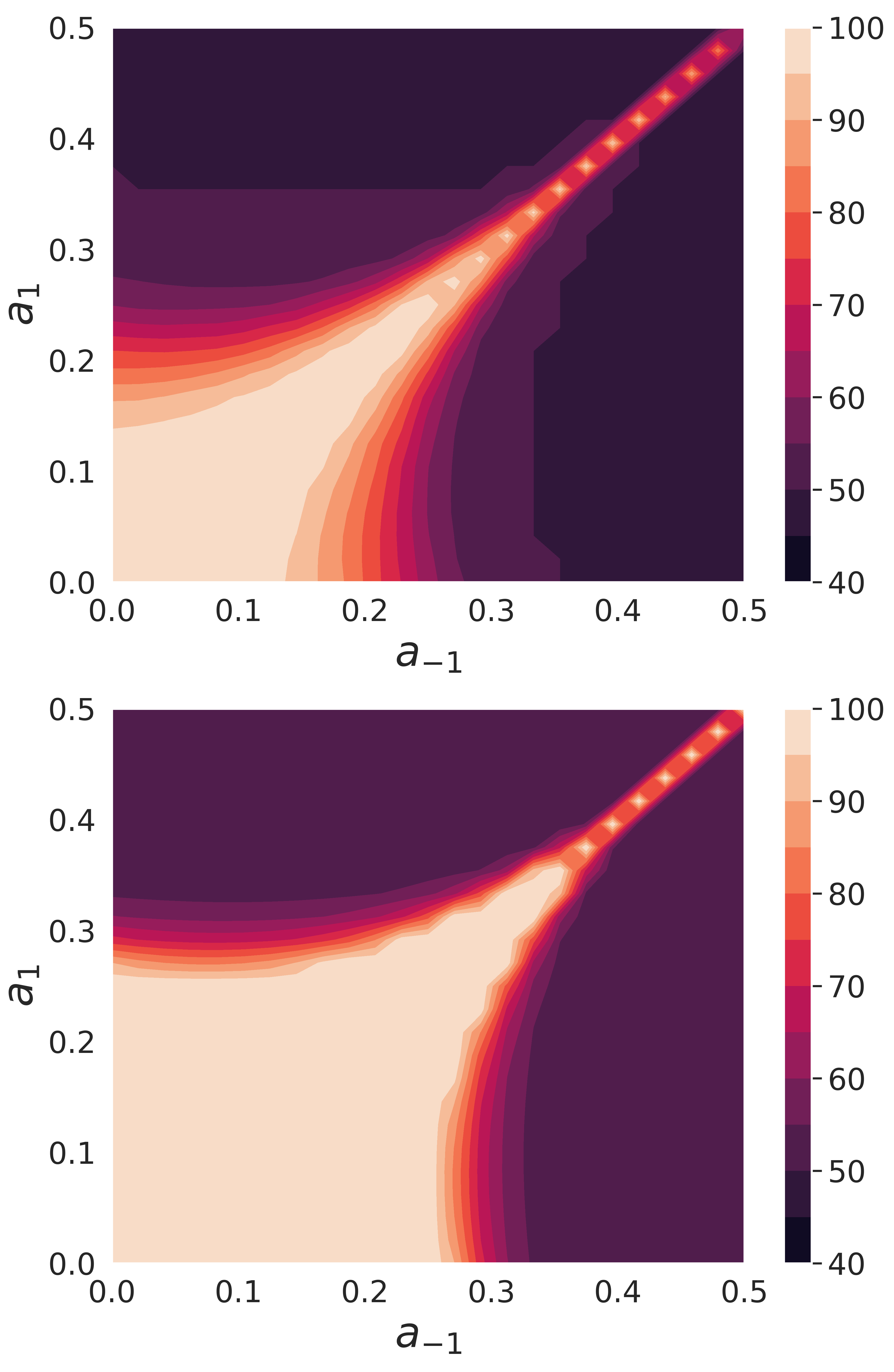}}
    \hspace{0.25cm}
  \subfloat[$\|\bm{\mu}\|_2 = 5.0$\label{fig:theory3}]{%
        \includegraphics[width=0.2\linewidth]{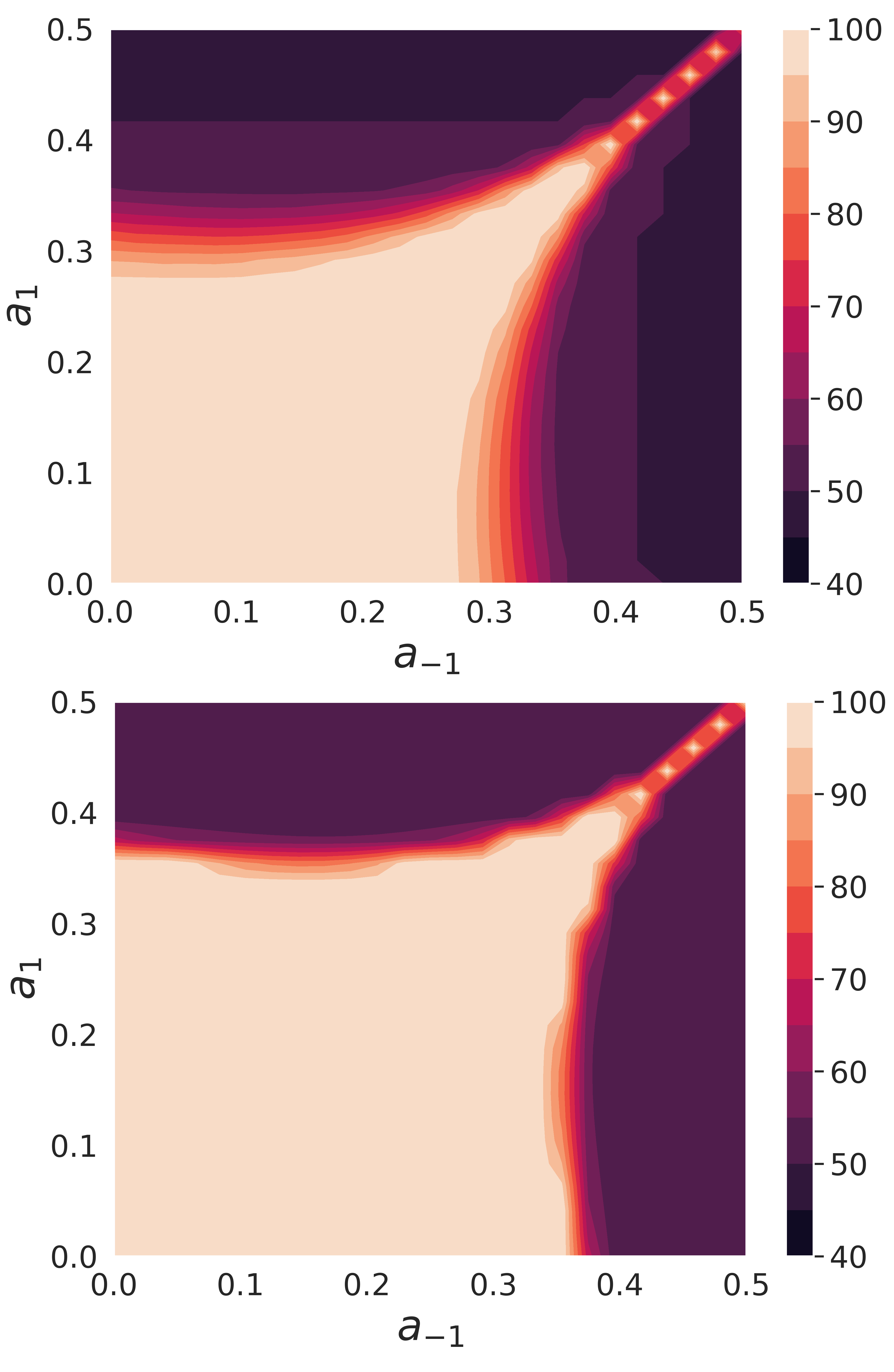}}

  \caption{We plot the contour plots for clean test accuracy of the CUDA Bayes classifier $\tilde{P}$ (given as $\tau_{\mathcal{D}}(\tilde{P})$) for varying $a_{y}$ parameters where $y \in \{\pm1\}$ in Figures \ref{fig:theory}(a--c). Gaussian mixture model $\mathcal{N}(y\bm{\mu}, I)$ denotes the clean data. In the top row, we use Lemmas \ref{lemma:bayesclean} \& \ref{lemma:bayesfun} to empirically generate the plots. In the bottom row, we plot the theoretical upper bound of $\tau_{\mathcal{D}}(\tilde{P})$ we obtain using Theorem \ref{theorem:funbound}.}
  \label{fig:theory}
\end{figure*}

In this section, we define a binary classification setup similar to \cite{at4, gmmsetup1} to theoretically analyze CUDA. Let $\mathcal{D}$ be a clean dataset modelled by an isotropic Gaussian mixture model given by $\mathcal{N}(y\bm{\mu}, I)$, where $y \in \{\pm1\}$ is the class label, $\bm{\mu} \in \mathbb{R}^d$, and $I \in \mathbb{R}^{d\times d}$ is the identity matrix. We defer the proofs for all the lemmas and theorem to the supplementary material. The Bayes optimal decision boundary for classifying this Gaussian mixture model is as follows.

\begin{lemma}
\label{lemma:bayesclean}
The Bayes optimal decision boundary for classifying $\mathcal{D}$ is given by $P(\bm{x}) \equiv \bm{\mu}^{\top} \bm{x} = 0$. The accuracy of the decision boundary $P$ on the clean dataset $\mathcal{D}$ (i.e.~$\tau_{\mathcal{D}}(P)$) is equal to $\phi(\|\bm{\mu}\|_2)$. Here, $\phi(.)$ represents the CDF of the standard normal distribution.
\end{lemma}
Now, to obtain $\mathcal{\tilde{D}}$ (i.e., CUDA), we perform class-wise 1D convolutions. Here, we consider a class of 1D convolutional filters with kernel size 3 of the form $\bm{f_a} = [a, 1, a]$ where $a \in [0, 0.5]$. The convolution of signal $\bm{x} \in \mathbb{R}^d$ with filter $\bm{f_a}$ using stride 1 (denoted by $\bm{x}*\bm{f_a}$), can be treated as a matrix operation $A\bm{x}$. Here, $A \in \mathbb{R}^{d \times d}$ is a tri-diagonal Toeplitz matrix denoted by $T(d; a, 1, a)$. For CUDA generation, we use class-wise convolution matrices $A_y = T(d; a_y, 1, a_y)$ to perturb a clean data point $(\bm{x}, y) \sim \mathcal{D}$ to an unlearnable data point $(A_y \bm{x}, y)$, where $a_y \in [0, 0.5]$ for $y \in \{\pm1\}$. Note that in CUDA, the labels remain the same. Next we show that such a perturbed dataset (i.e. CUDA) can be represented as a Gaussian mixture model.

\begin{lemma}
\label{lemma:perturbdistbn}
A CUDA generated from clean data distribution $\mathcal{D}$ and the $A_y$'s defined above can be modelled as a Gaussian mixture model with the distribution given by $\tilde{\mathcal{D}} = \mathcal{N}(y A_y \bm{\mu}, A_y^2)$. 
\end{lemma}

To characterize the decision boundary for CUDA, we need to use some properties of Toeplitz matrices from Noschese \textit{et al.} \cite{toeplitz} given in the following Lemma \ref{lemma:topelitz}.

\begin{lemma}
\label{lemma:topelitz}
 Any tri-diagonal Toeplitz matrix $A = T(d; a, 1, a) \in \mathbb{R}^{d\times d}$ with $a \in [0, 0.5]$ can be diagonalized as $A = Q D Q$ where $Q_{i,j} = \big({\frac{2}{d+1}}\big)^{1/2} \sin \big(\frac{ij\pi}{d+1}\big)$ and $D$ is a diagonal matrix with $D_{i,i} = 1 + 2a \cos \big(\frac{i\pi}{d+1}\big)$ for $1 \leq i,j \leq d$.
\end{lemma}
Next we use Lemma \ref{lemma:topelitz} to show that the Bayes optimal decision boundary for classifying $\tilde{\mathcal{D}}$ is a quadratic plane. 
\begin{lemma}
\label{lemma:bayesfun}
Let $A_{-1} = T(d; a_{-1}, 1, a_{-1})$ and $A_1 = T(d; a_1, 1, a_1)$. The Bayes optimal decision boundary for classifying $\tilde{\mathcal{D}}$ is given by $\tilde{P}(\bm{x}) \equiv  \bm{x}^{\top} A \bm{x} + \bm{b}^{\top} \bm{x}  + c = 0$, where $A = A_{-1}^{-2} - A_{1}^{-2},~ \bm{b} = 2(A_{-1}^{-1} + A_{1}^{-1})\bm{\mu}$, and $c = 2\sum_{i=1}^d [\ln (1 + 2 a_{-1} \cos (\frac{i \pi}{d+1})) - \ln (1 + 2 a_{1} \cos (\frac{i \pi}{d+1}))]$.
\end{lemma}
Now we state a lemma regarding the tail of Gaussian quadratic forms which plays a crucial role in our main result.
\begin{lemma}
\label{lemma:chernoff}
Let $\|\cdot\|$ denote the operator norm, $\|\cdot\|_2$ denote the vector 2-norm,  $\bm{z} \sim \mathcal{N}(\bm{0}, I)$, and $Z = \bm{z}^\top A \bm{z} + \bm{z} ^ \top \bm{b} + c$ where $A = Q \Lambda Q^\top$. Using Chernoff bound, for any $t \geq 0$ and $\gamma \in \mathbb{R}$,

\begin{align}
    \mathbb{P}\{Z - \mathbb{E} Z \geq \gamma \} &\leq \nonumber \\ &\frac{\exp\big\{\frac{-t}{4\|\bm{b}\|^2_2 \|\Lambda \|} -t (\gamma + \Tr(\Lambda) + \|\bm{b}\|_2) \big\}}{|I-2t\Lambda|^{\frac{1}{2}}}. \nonumber
\end{align}

\end{lemma}
Lemma \ref{lemma:chernoff} allows us to provide an upper bound for the accuracy of the unlearnable decision boundary $\tilde{P}$ on the clean dataset $\mathcal{{D}}$, given as $\tau_{\mathcal{D}}(\tilde{P})$, in Theorem \ref{theorem:funbound} below. 

\begin{theorem}[Main result]
\label{theorem:funbound}
Let $A = Q \Lambda Q = A^{-2}_{-1} - A^{-2}_{1}$. For any non-negative constants $t_1$ and  $t_2$, the accuracy of the unlearnable decision boundary $\tilde{P}$ on the clean dataset $\mathcal{{D}}$ can be upper-bounded as
\begin{align*}
    \tau_{\mathcal{D}}(\tilde{P}) &\leq \frac{\exp \big[t_1 \big( \bm{b}^\top \bm{\mu} + \bm{\mu}^\top A \bm{\mu} + c - \frac{1}{4\|2A\bm{\mu} + \bm{b}\|_2 \|\Lambda\|} \big) \big]}{2|I - 2t_1 \Lambda|^{\frac{1}{2}}} \\
    &+ \frac{\exp \big[t_2 \big( \bm{b}^\top \bm{\mu} -\bm{\mu}^\top A \bm{\mu} - c - \frac{1}{4\|2A\bm{\mu} - \bm{b}\|_2 \|\Lambda\|} \big) \big] }{2|I - 2t_2 \Lambda|^{\frac{1}{2}}}  
    \\&:=p_1 + p_2.
\end{align*} 
Furthermore, if $\bm{\mu}^\top A \bm{\mu} + \bm{b}^\top \bm{\mu} + c + \Tr(A) + \|2A\bm{\mu} + \bm{b}\| < 0$, we have $\tau_\mathcal{D}(\tilde{P}) \leq \frac{1}{2}(p_1 + 1) < 1$. Also, if  $-\bm{\mu}^\top A \bm{\mu} + \bm{b}^\top \bm{\mu} - c - \Tr(A) + \|2A\bm{\mu} - \bm{b}\| < 0$, we have $\tau_\mathcal{D}(\tilde{P}) \leq \frac{1}{2}(1 + p_2) < 1$. Moreover, for any $\bm{\mu} \neq \bm{0}$ and $a_{-1} \in [0, 0.5], ~ \exists a_1$ such that $\tau_{\mathcal{D}}(\tilde{P}) < \tau_{\mathcal{D}}({P})$.
\end{theorem}
Poisoning is \textit{effective} only if the accuracy of the unlearnable model $\tilde{P}$ is less than that of the clean model $P$ on the clean dataset $\mathcal{D}$, that is, $\tau_{\mathcal{D}}(\tilde{P}) < \tau_{\mathcal{D}}(P)$. To satisfy this condition, we need to carefully select $a_y$'s. In Theorem \ref{theorem:funbound}, we formally state this condition\footnote{We note that the conditions $\bm{\mu}^\top A \bm{\mu} + \bm{b}^\top \bm{\mu} + c + \Tr(A) + \|2A\bm{\mu} + \bm{b}\|_2 < 0$ or $-\bm{\mu}^\top A \bm{\mu} + \bm{b}^\top \bm{\mu} - c - \Tr(A) + \|2A\bm{\mu} - \bm{b}\|_2 < 0$ can always be satisfied by picking a sufficiently large $\bm{\mu}$ in the direction of an eigenvector corresponding to a negative or positive eigenvalue of $A$ (note that $A$ has negative and positive eigenvalues).}. Theorem \ref{theorem:funbound} shows that CUDA can effectively poison when there are two distinct modes in the clean Gaussian mixture data model. We validate our theoretical claim through empirical analysis as well (see Figure \ref{fig:theory}). Details for the analysis is given in supplementary material. 
We find that our upper bound for the clean test accuracy of CUDA classifiers are consistent with our empirical analysis. In our experiments, we also find that the unlearnability effect is stronger with a larger blur parameter. This effect is evident from Figure \ref{fig:theory} where we find that it is likely to get a lower $\tau_{\mathcal{D}}(\tilde{P})$ with higher $a_y$ values. These results are consistent with our experimental results in Section \ref{sec:experiments} with CIFAR-10, CIFAR-100, and ImageNet-100 datasets.
\section{Experiments} \label{sec:experiments}
In this section, we first discuss our experimental setup. More details on the setup is deferred to supplementary material. 
We then show the robustness of CUDA generation with various datasets and architectures. We also run various experiments to analyze the effectiveness of CUDA under different training techniques (ERM, AT with varying budgets, randomized smoothing, transfer learning, and fine-tuning) and augmentation techniques (mixup \cite{mixup}, grayscaling, random blurring, cutout \cite{cutout}, cutmix \cite{cutmix}, autoaugment \cite{autoaugment}, and  orthogonal regularization \cite{orth}). Finally, we also design adaptive defenses to test the robustness of CUDA. One might think that CUDA filters can be obtained by adversarially training them with the data. We show that CUDA is robust to such adaptive defenses that we design.

\begin{table}[t]

  \caption{Test accuracy ($\%$) of ResNet-18 trained on various unlearnable datasets. We use $L_{\infty}$ AT with budget $\rho_a = 4/255$. }
  \label{tab:datasets}
  \scriptsize
  \centering 
  \begin{tabular}{p{1.4cm} p{0.7cm} p{0.5cm} p{0.4cm} p{0.4cm} p{0.4cm} p{0.4cm} p{0.5cm}} 
    \toprule
    \multirow{2}{1.4cm}{\centering Dataset} & \multirow{2}{0.7cm}{\centering Training method} & \multirow{2}{0.5cm}{\centering Clean} & \multicolumn{5}{c}{Unlearnability method} \\
    \cmidrule(l){4-8}
    {} & {} & {} & EM & TAP & NTGA & REM & \textbf{CUDA} \\
    \toprule
    \multirow{2}{1.4cm}{\centering CIFAR-10} & ERM & 94.66 & \textbf{13.20} & 22.51 & 16.27 & 27.09 & 18.48\\ 
         {} & AT & 89.51 & 88.62 & 88.02 & 88.96 & 48.16 & \textbf{44.40}\\
         \cmidrule[0.1ex]{1-8}
         \multirow{2}{1.5cm}{\centering CIFAR-100} & ERM & 76.27 & \textbf{1.60} & 13.75 & 3.22 & 10.14 & 12.69\\
         {} & AT & 64.50 & 63.43 & 62.39 & 62.44 & \textbf{27.10} & 34.34\\
         \cmidrule[0.1ex]{1-8}
         \multirow{2}{1.5cm}{\centering ImageNet-100} & ERM & 80.66 & \textbf{1.26} & 9.10 & 8.42 & 13.74 & 8.96\\ 
         {} & AT & 66.62 & 63.40 & 63.56 & 63.06 & 41.66 & \textbf{38.68}\\
    \bottomrule[0.25ex]
  \end{tabular}
\end{table}

\subsection{Experimental setup}\label{sec:expsetup}

\textbf{Datasets.} We use three image classification datasets -- CIFAR-10, CIFAR-100, \cite{cifar} and ImageNet-100 (a subset of ImageNet made of the first 100 classes) \cite{imagenet}. We use the data augmentation techniques such as random flipping, cropping, and resizing \cite{augment}.

\textbf{Architectures.} We use ResNet-18 \cite{resnet}, VGG-16 \cite{vgg}, Wide ResNet-34-10 \cite{wrn}, and DenseNet-121 \cite{densenet}. We train the networks with hyperparameters used in Fu \textit{et al.} \cite{rem}. Previous works mainly employ ResNet-18 for most of their evaluations. Additional experiments on Tiny-ImageNet \cite{le2015tiny}, DeIT \cite{deit}, EfficientNetV2-S \cite{en}, and MobileNetV2 \cite{mn} are provided in the supplementary material.

\textbf{CUDA generation.} We use filters of size $k=3$ and blur parameter $p_b=0.3$ for both CIFAR-10 and CIFAR-100 datsets. ImageNet-100 is a higher dimensional 224$\times$224$\times$3  image dataset when compared to the 32$\times$32$\times$3 dimensional CIFAR datasets. Hence, we use larger filters of size $k=9$ with $p_b=0.06$ for ImageNet-100. These hyperparameters are chosen such that the CUDA images are not perceptibly highly perturbed and give good unlearnability effect (see plot in Figure \ref{fig:funcifar}). 
In supplementary matrial, 
we show the results of training on CUDA with different hyperparameters for data generation. 

\textbf{Baselines.} We compare CUDA generation technique with four state-of-the-art unlearnability methods -- REM \cite{rem}, EM \cite{em}, TAP \cite{tap}, and NTGA \cite{ntga}. We adopt the results reported in \cite{rem} since we use the same hyperparameters for training. For REM, we select hyperparameters $\rho_u = 8/255$ and $\rho_a = 4/255$, the highest radii values in \cite{rem}. For comparing unlearnable methods, we look at the clean test accuracy. The lower the test accuracy, the better the unlearnability method is. As mentioned in the supplementary material, we use publicly released official codebases for reproducing the baselines using their default hyperparameters.

\begin{table}[t]
  \caption{Test accuracy ($\%$) of network architectures trained on various CIFAR-10 unlearnable datasets with $L_{\infty}$ AT ($\rho_a = 4/255$).}
  \label{tab:archs}
  \scriptsize
  \centering
  \begin{tabular}{c c p{0.4cm} p{0.4cm} p{0.4cm} p{0.4cm} p{0.5cm}}
    \toprule
    \multirow{2}{2cm}{\centering Model} & \multirow{2}{1cm}{\centering Clean} & \multicolumn{5}{c}{Unlearnability method} \\
    \cmidrule(l){3-7}
    {} & {} & EM & TAP & NTGA & REM & \textbf{CUDA}\\
    \toprule
    ResNet-18 & 89.51 & 88.62 & 88.02 & 88.96 & 48.16 & \textbf{44.40}\\
    \cmidrule[0.1ex]{1-7}
    VGG-16 & 87.51 & 86.48 & 86.27 & 86.65 & 65.23 & \textbf{42.98}\\
    \cmidrule[0.1ex]{1-7}
    Wide ResNet-34-10 & 91.21 & 90.05 & 90.23 & 89.95 & \textbf{48.39} & 53.02\\
    \cmidrule[0.1ex]{1-7}
    DenseNet-121 & 83.27 & 82.44 & 81.72 & 80.73 & 81.48 & \textbf{45.95}\\
    
    \bottomrule[0.25ex]
  \end{tabular}
\end{table}

\begin{table}[t]
  \caption{Test accuracy ($\%$) of ResNet-18 with CUDA under various training settings.}
  \label{tab:robustfun}
  \scriptsize
  \centering
  \begin{tabular}{c c l c}
    \toprule
    Dataset & Clean & Training method & \textbf{CUDA (ours)} \\
    \midrule
    \multirow{7}{1.4cm}{\centering CIFAR-10} & \multirow{7}{1cm}{\centering 94.66} 
    & ERM & 18.48\\
    {} & {}    &AT $L_{\infty}~(\rho_a = 4/255)$  & $44.40$\\
    {} & {}      & AT $L_{\infty}~(\rho_a = 8/255)$  & 32.85\\
    {} & {}    & AT $L_{\infty}~(\rho_a = 16/255)$  & 19.32\\
    {} & {}     & AT $L_{2}~(\rho_a = 0.25)$  & 39.05\\
    {} & {}     & AT $L_{2}~(\rho_a = 0.50)$  & 51.19\\
    {} & {}     & AT $L_{2}~(\rho_a = 0.75)$  & 51.14\\
    \cmidrule{1-4}
    \multirow{4}{1.4cm}{\centering CIFAR-100} & \multirow{4}{1cm}{\centering 76.27} 
    & ERM & 12.69\\
    {} & {}   &AT $L_{\infty}~(\rho_a = 4/255)$  & 34.34\\
    {} & {}    & AT $L_{\infty}~(\rho_a = 8/255)$  & 30.00\\
    {} & {}     & AT $L_{2}~(\rho_a = 0.75)$  & 36.90\\
    \cmidrule{1-4}
    \multirow{4}{1.5cm}{\centering ImageNet-100} & \multirow{4}{1cm}{\centering 80.66} & ERM & 8.96\\
    {} & {}  &AT $L_{\infty}~(\rho_a = 4/255)$  & 38.68\\
    {} & {} & AT $L_{\infty}~(\rho_a = 8/255)$  & 40.08\\
    {} & {}     & AT $L_{2}~(\rho_a = 0.75)$  & 20.58\\

    \bottomrule[0.25ex]
  \end{tabular}
\end{table}

\begin{table*}[t]
\centering
\caption{Test accuracy $(\%)$ of ResNet-18 on CIFAR-10 with different data protecion percentages. The last row shows the results for CUDA with ERM setting. The rest of the rows show results for unlearnability methods trained in the $L_{\infty}$ AT ($\rho_a = 4/255$) setting.
}
\label{tab:mix}
\footnotesize
\begin{tabular}{c c | c c | c c | c c | c c | c}
\toprule
 \multirow{3}{2cm}{\centering Unlearnability method} & \multicolumn{10}{c}{\centering Data Protection Percentage} \\
\cmidrule{2-11}
 & \multirow{2}{0.4cm}{\centering 0\%} & \multicolumn{2}{c|}{\centering 20\%} & \multicolumn{2}{c|}{\centering 40\%} & \multicolumn{2}{c|}{\centering 60\%} & \multicolumn{2}{c|}{\centering 80\%} & \multirow{2}{0.4cm}{\centering 100\%} \\
 & & Mixed & Clean & Mixed & Clean & Mixed & Clean & Mixed & Clean & \\
\midrule
 EM & \multirow{4}{0.5cm}{\centering 89.51} & 89.60 & \multirow{4}{0.5cm}{\centering 88.17} & 89.40 & \multirow{4}{0.5cm}{\centering 86.76} & 89.49 & \multirow{4}{0.5cm}{\centering 85.07} & 89.10 & \multirow{4}{0.5cm}{\centering 79.41} & 88.62 \\
TAP &
& 89.01 && 88.66 && 88.40 && 88.04 && 88.02 \\
NTGA &
& 89.56 && 89.35 && 89.22 && 89.17 && 88.96 \\
REM &
& 89.60 && 89.34 && 89.61 && 88.09 && 48.16 \\
\textbf{CUDA (ours)} &
& \textbf{88.54} && \textbf{87.24} && \textbf{86.03} && \textbf{84.34} && \textbf{44.40} \\
\midrule
\textbf{CUDA + ERM} & 94.66 & 93.28 & 93.75 & 91.34 & 92.56 & 89.91 & 89.77 & 85.61 & 84.30 & 18.48\\
\bottomrule
\end{tabular}
\end{table*}

\subsection{Effectiveness of CUDA}
\label{sec:effect_fun}

\textbf{Why does CUDA work?} In order to measure how much CUDA technique's blurring affects the dataset's quality, we compare class-wise blurring (CUDA technique) against universal blurring where a single convolutional filter is used for blurring all the images. We keep the filter generation parameters fixed ($p_b=0.3$ and $k=3$) for both CUDA class-wise blurring and universal blurring. ResNet-18 trained with clean CIFAR-10, universally blurred CIFAR-10, and CIFAR-10 CUDA achieve clean test accuracies of 94.66$\%$, 90.47$\%$, and 18.48$\%$, respectively. This suggests that our controlled blurring does not obscure the semantics of the dataset. Hence, the significant drop in the clean test accuracy introduced by CUDA is most likely due to the \textit{usage of class-wise filters}. This suggests that a model trained on CUDA learns the relation between the class-wise convolution filters and their corresponding labels. Therefore, during test time when this convolution effect is absent, the CUDA trained model fails to make correct classifications. Furthermore, the model trained on CUDA achieves an accuracy of 99.91$\%$ on the CIFAR-10 CUDA testset. This strongly supports our claim that the CUDA model learns to classify images based on the convolutional filters used to blur them. In addition, if we permute the class-wise filters for blurring the test set (i.e., blurring class 1 images with class 2 filters, class 2 images with class 3 filters, and so on), we get a very low  accuracy of 2.53$\%$ on this test set. Further details are provided in the supplementary material. 
Finally, we note that real-world datasets might also contain blurred images due to various factors such as motion blurring, weather conditions, issues with the camera, etc. Hence, detecting if a blurred image is poisoned might not always be possible. However, one might argue that it is possible to detect if an entire dataset is blurred. Interestingly, later in this section we show that CUDA technique exhibits unlearnability effect even when only a fraction of the training dataset is poisoned.

\textbf{Different datasets.} We first compare the effectiveness of CUDA with different datasets using ERM and $L_{\infty}$ AT with $\rho_a = 4/255$. We use ResNet-18 for the experiments. The results are shown in Table \ref{tab:datasets}. These results show that EM, TAP, and NTGA are not robust to AT. However, both CUDA and REM are successful. Here, our method CUDA outperforms REM with CIFAR-10 and ImageNet-100 datasets. Smartly designed additive noise in AT helps in achieving better generalization than ERM on the unlearnable datasets. This experiment thus demonstrates that ERM and AT are not good choices for training with CUDA and REM dataset.

\textbf{Different models.} Next we compare the effectiveness of CUDA using various deep learning architectures with $L_{\infty}$ AT ($\rho_a = 4/255$). We use CIFAR-10 for the experiments. The results are shown in Table \ref{tab:archs}. As we see in the table, CUDA is effective with all the five network architectures. However, REM is not seen to be transferrable with DenseNet-121.

\textbf{Robustness to different training settings.} In Section \ref{sec:limitations}, we show that REM is sensitive to the training settings. REM generated using $L_{\infty}$ radii budgets of $\rho_u = 8/255$ and $\rho_a = 4/255$ for CIFAR-10 breaks with $L_{\infty}$ AT ($\rho_a = 8/255$) and $L_2$ AT ($\rho_a = 0.75$) to get test accuracy of 78.71$\%$ and 79.65$\%$, respectively. Hence, we run experiments to check the robustness of CUDA with various AT norm budgets. The results are shown in Table \ref{tab:robustfun}. As we see in the table, CUDA is robust to ERM,  $L_{\infty}$, and $L_2$ AT settings with varying training budgets. Impressively, the highest test accuracy achieved with training on CIFAR-10 CUDA, CIFAR-100 CUDA, and ImageNet-100 CUDA are as low as 51.19$\%$, 36.90$\%$, and 40.08$\%$, respectively. We also find that using a pre-trained ResNet-18 with CIFAR-10 CUDA only achieves clean test accuracy of 42.42$\%$ and 48.22$\%$ with fine-tuning the full network and a newly trained final layer, respectively (details are deferred to the supplementary material).

\textbf{Different protection percentages.} In Section \ref{sec:preliminaries}, we assume that the attacker has access to the full clean training data. However, in real life settings, this might not be always possible. Hence, we train ResNet-18 on a mix of CIFAR-10 CUDA and clean CIFAR-10 training datasets to evaluate the effectiveness of poisoning with varying data protection percentages. Protection percentage denotes the percentage of the training data that is poisoned.

We show the results in Table \ref{tab:mix}. In the table, the ``Mixed'' column denotes the clean test accuracy of a model trained using a mix of both clean and poisoned data. The ``Clean'' column denotes the clean test accuracy of a model trained only using the clean subset of the training data. In the last row of Table \ref{tab:mix}, we provide the results for CUDA trained using ERM with varying data protection percentages. The remainder of the rows provide the results for the $L_{\infty}$ AT ($\rho_a = 4/255$) scenario. For example, CUDA trained with ERM using an 80$\%$ clean training data partition achieves a test accuracy of 93.75$\%$. Adding the 20$\%$ CUDA data partition to the training dataset drops the test accuracy of the model to 93.28$\%$. Results from the last row of Table \ref{tab:mix} show that CUDA with varying data protection percentages is effective with the ERM setting. However, with the AT scenario, the unlearnability techniques are not as successful as with ERM with varying data protection percentages. 
Nevertheless, it is interesting to note that CUDA technique performs better than all the other unlearnability techniques with AT.

\textbf{Robustness to smoothing and data augmentations.} Cohen \textit{et al.} \cite{smoothingcohen} proposes randomized smoothing which is a provable adversarial defense  in $L_2$ norm. 
Since CUDA does not use additive noise, CUDA is robust to randomized smoothing. A smoothed ResNet-18 (with a noise level of 0.5) with CIFAR-10 CUDA achieves only a clean test accuracy of 43.85$\%$. In Section \ref{sec:limitations}, we see that REM breaks with grayscaling. While ResNet-18 training using grayscaled REM achieves 70.76$\%$ test accuracy, grayscaled CUDA only achieves 20.12$\%$ test accuracy on the clean grayscaled CIFAR-10 test data. This shows that CUDA technique exhibits the desirable property of not relying upon the color space for its attack. CIFAR-10 CUDA training achieves only 25.53$\%$, 25.80$\%$, 26.93$\%$, 34.09$\%$, and 50.72$\%$ clean test accuracy with mixup, cutout, cutmix, autoaugment, and orthogonal regularization, respectively (see supplementary material for more details).

\textbf{Adaptive defenses for CUDA.} Here, we first investigate the effect of training CIFAR-10 CUDA with random blurring augmentations using ResNet-18. Each batch of the CUDA training data is convolved with random $3\times 3$ filters of varying blur parameters $p_b'$. With $p_b'$ values of 0.1 and 0.3, CUDA training achieves lower test accuracy 9.2$\%$ and 13.37$\%$, respectively. This shows that the noise added by CUDA technique is robust to random convolution augmentations. 

One may think that CUDA technique can be broken by learning the private filters from the data. We test this idea by training deconvolution filters to find if we can reverse the blurring effect in CUDA with adversarially trained filters. We use a novel Deconvolution-based Adversarial Training (DAT) technique that is similar to AT (check supplementary material for details). 
While the adversarial step in AT learns sample-wise error-maximizing additive noises, the adversarial step in DAT learns class-wise error-maximizing deconvolution filters. We train DAT with filters of varying sizes (3, 5, and 7) on CIFAR-10 CUDA using ResNet-18. The filter parameters are constrained within a finite range to make sure that the images do not get distorted with the adversarial step similar to projection in projected gradient descent.  We find that CUDA is robust to DAT. DAT using filters of size 3, 5, and 7 with CUDA achieves only test accuracy of 39.05$\%$, 46.21$\%$, and 38.48$\%$, respectively. DAT is not successful against CUDA since we can not invert convolutions without the knowledge of the private filters or clean images corresponding to CUDA.

\textbf{Limitations and future directions.} The unlearnability effect of CUDA can be defended if some fraction of the clean data and its corresponding CUDA images are leaked. The defender must also be able to detect if all samples in the dataset are poisoned. However, our work assumes a setup where the filters remain private. For example, a data publisher could simply publish their CUDA images and delete the clean images permanently to prevent this scenario. As discussed in this section, CUDA as well as other prior works do not perform well with different protection percentages with AT. Improving this can be an interesting research direction. We believe that extending CUDA technique to other domains such as tabular and text data is also an interesting future direction. It would also be interesting to see theoretical analysis of CUDA considering more complex setups.

\section*{Acknowledgement}
This project was supported in part by Meta grant 23010098, NSF CAREER AWARD 1942230, HR001119S0026 (GARD), ONR YIP award N00014-22-1-2271, Army Grant No. W911NF2120076, a capital one grant, NIST 60NANB20D134 and the NSF award CCF2212458.

{\small
\bibliography{references}
\bibliographystyle{ieee_fullname}
}

\appendix
\section{Appendix}

\subsection{Proof for Lemma \ref{lemma:bayesclean}}
\label{proof:bayesclean}

At the optimal decision boundary the probabilities of any point $\bm{x} \in \mathbb{R}^d$ belonging to class $y=-1$ and $y=1$ modeled by $\mathcal{{D}}$ are the same. Here, $\bm{\mu} = \bm{\mu_{1}} = -\bm{\mu_{-1}}$ and $I = \Sigma_{-1} = \Sigma_1$.

\begin{align*}
    &\implies \dfrac{\exp[-\frac{1}{2} (\bm{x} - \bm{\mu_{-1}})^\top \Sigma_{-1}^{-1} (\bm{x} - \bm{\mu_{-1}})]}{\sqrt{(2 \pi)^d |\Sigma_{-1}|}}  =\\ &\dfrac{\exp[-\frac{1}{2} (\bm{x} - \bm{\mu_1})^\top \Sigma_1^{-1} (\bm{x} - \bm{\mu_1})]}{\sqrt{(2 \pi)^d |\Sigma_1|}} \\
&\implies -\frac{1}{2}\log |I| - \frac{1}{2} (\bm{x}^\top \bm{x} - 2 \bm{x}^\top \bm{\mu_{-1}} + \bm{\mu_{-1}}^\top \bm{\mu_{-1}}) =\\ &-\frac{1}{2}\log |I| - \frac{1}{2} (\bm{x}^\top \bm{x} - 2 \bm{x}^\top \bm{\mu_1} + \bm{\mu_1}^\top \bm{\mu_1})\\
&\implies \bm{x}^\top (\bm{\mu_{-1}} - \bm{\mu_1}) - \frac{1}{2} (\bm{\mu_{-1}}^\top \bm{\mu_{-1}} - \bm{\mu_1}^\top \bm{\mu_1}) = 0\\
&\implies -\bm{x}^\top \bm{\mu} = 0\\
&\implies P(\bm{x}) \equiv \bm{x}^\top \bm{\mu} = 0.
\end{align*}

Now, the accuracy of the clean model $P$ is to be computed. Note that if $P(\bm{x}) < 0$ the Bayes optimal classification is class -1, else the classification is class 1. Let $\bm{z} \sim \mathcal{N}(\bm{0}, I)$, and $Z \sim \mathcal{N}(0, 1)$, and $sgn(.)$ be the signum function.

\begin{align*}
    \tau_{\mathcal{D}}(P) &= \mathbb{E}_{(\bm{x},y)\sim \mathcal{D}}~[\mathbbm{1}(y = sgn(P(\bm{x})))] =  \mathbb{P}[y \bm{x}^\top \bm{\mu} > 0] \\&= \mathbb{P}[y (y\bm{\mu} + \bm{z})^\top \bm{\mu} > 0]\\
    &= \mathbb{P}[(\bm{\mu} + \bm{z})^\top \bm{\mu} > 0]\\
    &= \mathbb{P}[\|\bm{\mu}\|_2^2 + \|\bm{\mu}\|_2 Z > 0] = \phi(\|\bm{\mu}\|_2).
\end{align*}  \hfill $\square$

\subsection{Proof for Lemma \ref{lemma:perturbdistbn}}
\label{proof:perturbdistbn}

Let $\mathcal{D}_1 = \mathcal{N}(\bm{\mu}, I)$. For every data point $(\bm{x}, y) \sim \mathcal{D}_1$, let the perturbed data $(A_1\bm{x}, y)$ be modelled by a distribution $\tilde{D}_1$. We prove that $\tilde{D}_1 = \mathcal{N}(A_1 \bm{\mu}, A_1^\top A_1)$. 
\begin{align*}
    \mathbb{E}_{(\bm{x}, y) \sim \mathcal{D}_1}~ A_1 \bm{x} = A_1 \mathbb{E}_{(\bm{x}, y) \sim \mathcal{D}_1}~ \bm{x}  = A_1 \bm{\mu}.
\end{align*}
\begin{align*}
    &\mathbb{E}_{(\bm{x}, y) \sim \mathcal{D}_1}~ (A_1 \bm{x} - A_1 \bm{\mu}) (A_1 \bm{x} - A_1 \bm{\mu})^\top \\&= \mathbb{E}_{(\bm{x}, y) \sim \mathcal{D}_1}~ A_1( \bm{x} -  \bm{\mu}) [A_1 (\bm{x} -  \bm{\mu})]^\top\\
    &= \mathbb{E}_{(\bm{x}, y) \sim \mathcal{D}_1}~ A_1( \bm{x} -  \bm{\mu}) (\bm{x} -  \bm{\mu})^\top A_1^\top \\
    &= A_1 \mathbb{E}_{(\bm{x}, y) \sim \mathcal{D}_1}~ ( \bm{x} -  \bm{\mu}) (\bm{x} -  \bm{\mu})^\top A_1^\top\\
    &= A_1 I A_1^\top = A_1 A_1^\top.
\end{align*}

Tri-diagonal Toeplitz matrices $A_y = T(d; a_y, 1, a_y)$ are symmetric. Hence, $\mathcal{\tilde{D}} = \mathcal{N}(yA_y \bm{\mu}, A_y^2)$. \hfill $\square$

\subsection{Remarks on Lemma \ref{lemma:topelitz}}
\label{proof:toeplitz}

A tri-diagonal Toeplitz matrix $T(d; a_1, a_2, a_3)$ is represented as 

$$\begin{bmatrix}
a_2 & a_3 & 0 & 0 &  \dots & 0\\
a_1 & a_2 & a_3 & 0 &  \dots & 0 \\
0 & a_1 & a_2 & a_3 &  \dots & 0\\
\vdots & \vdots & \vdots & \vdots & \ddots & \vdots \\
0 & \dots & \dots &0 &a_1 & a_2
\end{bmatrix} \in \mathbb{R}^{d\times d}$$

The class of matrices $A_y = T(d; a_y, 1, a_y)$ are symmetric and can be diagonalized as $Q D Q^\top$. $Q = \big(\big({\frac{2}{d+1}}\big)^{1/2} \sin \big(\frac{ij\pi}{d+1}\big)\big)_{i,j}$ is symmetric and it is the common eigenvector matrix to all $A_y$ matrices. As shown in Lemma \ref{lemma:topelitz}, $Q$ and $D$ can be represented using trigonometric functions. Also, we have $A(n) := A_1^{n} \pm A_{-1}^{n} = Q (D_1^n \pm D_{-1}^n) Q$ where $A_1 = QD_1Q$ and $A_{-1} = QD_{-1}Q$. Further, $\Tr(A(n)) = \Tr( Q (D_1^n \pm D_{-1}^n) Q) = \Tr((D_1^n \pm D_{-1}^n) Q^2) = \Tr(D_1^n \pm D_{-1}^n)$.

\subsection{Proof for Lemma \ref{lemma:bayesfun}}
\label{proof:bayesfun}

At the optimal decision boundary the probabilities of any point $\bm{x} \in \mathbb{R}^d$ belonging to class $y=-1$ and $y=1$ modeled by $\mathcal{\tilde{D}}$ are the same. Here, $\bm{\mu} = \bm{\mu_{1}} = -\bm{\mu_{-1}}$ and $A_y$'s are symmetric.

\begin{align*}
    &\dfrac{\exp[-\frac{1}{2} (\bm{x} - A_{-1} \bm{\mu_{-1}})^\top (A_{-1} I A_{-1}^\top)^{-1} (\bm{x} - A_{-1} \bm{\mu_{-1}})]}{\sqrt{(2 \pi)^d |A_{-1} I A_{-1}^\top|}}  \\ &=\dfrac{\exp[-\frac{1}{2} (\bm{x} - A_1 \bm{\mu_1})^\top (A_1 I A_1^\top)^{-1} (\bm{x} - A_1 \bm{\mu_1})]}{\sqrt{(2 \pi)^d |A_1 I A_1^\top|}}\\
    &\implies -\frac{1}{2} \ln \frac{|A_{-1}^2|}{|A_1^2|} - \frac{1}{2} [\bm{x}^\top (A_{-1}^{-2} - A_1^{-2}) \bm{x}\\ &- 2(\bm{\mu_{-1}}^\top A_{-1}^{-1} - \bm{\mu_1}^\top A_1^{-1}) \bm{x}\\
    &+ (\bm{\mu_{-1}}^\top \bm{\mu_{-1}} - \bm{\mu_1}^\top \bm{\mu_1})] = 0\\
    &\implies \tilde{P}(\bm{x}) \equiv  \bm{x}^\top (A_{-1}^{-2} - A_1^{-2})\bm{x}\\ &- 2(\bm{\mu_{-1}}^\top A_{-1}^{-1} - \bm{\mu_1}^\top A_1^{-1}) \bm{x} +  (\|\bm{\mu_{-1}}\|_2^2 - \|\bm{\mu_1}\|_2^2)\\ 
    &+ \sum_{i=1}^d \ln \bigg( \dfrac{1 + 2 a_{-1} \cos (\frac{i \pi}{d+1})}{1 + 2 a_1 \cos (\frac{i \pi}{d+1})} \bigg)^2 = 0\\
    &\implies \tilde{P}(\bm{x}) \equiv  \bm{x}^\top (A_{-1}^{-2} - A_1^{-2})\bm{x} \\&+ 2[(A_{-1}^{-1} + A_1^{-1}) \bm{\mu}]^\top \bm{x} \\&+ \sum_{i=1}^d \ln \bigg( \dfrac{1 + 2 a_{-1} \cos (\frac{i \pi}{d+1})}{1 + 2 a_1 \cos (\frac{i \pi}{d+1})} \bigg)^2 = 0\\
     &\implies \tilde{P}(\bm{x}) \equiv \bm{x}^\top A \bm{x} + \bm{b}^\top \bm{x} + c = 0.
\end{align*}\hfill $\square$

Note that here if $\tilde{P}(\bm{x}) < 0$, the Bayes optimal classification is class -1, else the classification is class 1. Here, for shorthand notations we denote $A = (A_{-1}^{-2} - A_1^{-2}), ~\bm{b} =  2(A_{-1}^{-1} + A_1^{-1}) \bm{\mu}, ~ c = \sum_{i=1}^d \ln \bigg( \dfrac{1 + 2 a_{-1} \cos (\frac{i \pi}{d+1})}{1 + 2 a_1 \cos (\frac{i \pi}{d+1})} \bigg)^2$.

\subsection{Proof for Lemma \ref{lemma:chernoff}}
\label{proof:chernoff}

Let $Z = \bm{z}^\top A \bm{z} + \bm{z}^\top \bm{b} + c$ and $\bm{z} \sim \mathcal{N}(\bm{0}, I) \subset \mathbb{R}^d$ where $A = Q \Lambda Q^\top$. Also,
\begin{align*}
    &Z = \bm{z}^\top A \bm{z} + \bm{z}^\top \bm{b} + c \\&= \bigg(\bm{z} + \frac{1}{2}A^{-1}\bm{b}\bigg)^\top A \bigg(\bm{z} + \frac{1}{2}A^{-1}\bm{b}\bigg) + c - \frac{1}{4}\bm{b}^\top A^{-1} \bm{b}.
\end{align*}

For any $t\geq 0$ and $\bm{x} \sim \mathcal{N}(\bm{0}, I)$, we write the moment generating function for a quadratic random variable $Y = \bm{x}^\top A \bm{x}$ as \footnote{\cite{mgf}}

\begin{align*}
    &\mathbb{E}[\exp(tY)] = \frac{1}{(2\pi)^{d/2}} \int_{\mathbb{R}^d} \exp\{t \bm{x}^\top A \bm{x}\} \\ &\exp \Big\{ -\frac{1}{2}(\bm{x}-\bm{\mu})^\top (\bm{x}-\bm{\mu})\Big\} d\bm{x}\\
    &= \frac{\exp\{ -\bm{\mu}^\top \bm{\mu}/2\}}{(2\pi)^{d/2}} \int_{\mathbb{R}^d} \exp \Big\{ -\frac{1}{2} \bm{x}^\top (I - 2tA) \bm{x} + \bm{\mu}^\top \bm{x} \Big\} d\bm{x}\\
    &= \frac{\exp\{ -\bm{\mu}^\top \bm{\mu}/2\}}{(2\pi)^{d/2}}   \frac{(2\pi)^{d/2} \exp \Big\{ \frac{1}{2} \bm{\mu}^\top (I-2tA)^{-1} \bm{\mu} \Big\}}{|I-2tA|^{1/2}}\\
    &= \frac{ \exp \Big\{ -\frac{1}{2} \bm{\mu}^\top [I - (I-2tA)^{-1}] \bm{\mu} \Big\}}{|I-2tA|^{1/2}}.\\
    &\implies \mathbb{E}[\exp(tZ)]= \\ &\frac{\exp \{-\frac{\bm{b}^\top}{8} A^{-1} [I-(I-2tA)^{-1}]A^{-1}\bm{b} + t[c-\frac{\bm{b}^\top}{4}\ A^{-1} \bm{b}] \}}{|I-2tA|^{\frac{1}{2}}}.
\end{align*}

Using the Chernoff bound and $\mathbb{E} ~\bm{z}^{\top} A \bm{z} = \Tr(A \mathbb{E}[z z^\top]) = \Tr(A)$, for some $\gamma$,

\begin{align*}
    &\mathbb{P}\{ Z \geq \mathbb{E}[Z] + \gamma \} \leq \frac{\mathbb{E}[\exp(tZ)]}{\exp\{t[\gamma + \mathbb{E}(Z)]\}} =\\ &\frac{\exp \{-\frac{\bm{b}^\top}{8} A^{-1} [I-(I-2tA)^{-1}]A^{-1}\bm{b} + t[c-\frac{\bm{b}^\top}{4} A^{-1} \bm{b}] \}}{\exp\{t([\gamma + \Tr(A) + \|b\|_2 + c]\} |I-2tA|^{\frac{1}{2}}}.
\end{align*}

Let us take $\bm{u} = Q^\top \bm{b}$. Also, $-\Lambda^{-1}[I - (I-2t\Lambda)^{-1}]\Lambda^{-1} = 2t\Lambda^{-1}(I-2t\Lambda)^{-1}$ since $\Lambda$ is a diagonal matrix. Using Woodbury matrix identity, we get $(I-2t\Lambda)^{-1} = I - (I - \frac{1}{2t}\Lambda^{-1})^{-1}$. This gives us

\begin{align*}
    &\mathbb{P}\{ Z \geq \mathbb{E}[Z] + \gamma \} \leq \\ &\exp \{-\frac{1}{8}\bm{b}^\top A^{-1} [I-(I-2tA)^{-1}]A^{-1}\bm{b} \\&+ t[c-\frac{1}{4}\bm{b}^\top A^{-1} \bm{b}] \} \exp\{-t[\gamma + \Tr(A) + \|b\|_2 + c]\} \\ &|I-2tA|^{\frac{-1}{2}} \\
    &= \exp \{-\frac{1}{8}\bm{u}^\top \Lambda^{-1} [I-(I-2t\Lambda)^{-1}] \Lambda^{-1} \bm{u} \\&+ t[c-\frac{1}{4}\bm{u}^\top \Lambda^{-1} \bm{u}] \} \exp\{-t[\gamma + \Tr(\Lambda) + \|b\|_2 + c]\} \\ &|I-2t\Lambda|^{\frac{-1}{2}}\\
    &= \exp \{\frac{t}{4}\bm{u}^\top \Lambda^{-1}(I-2t\Lambda)^{-1} \bm{u} \\&+ t[c-\frac{1}{4}\bm{u}^\top \Lambda^{-1} \bm{u}] \} \exp\{-t[\gamma + \Tr(\Lambda) + \|b\|_2 + c]\} \\ &|I-2t\Lambda|^{\frac{-1}{2}}\\
    &= \exp \{\frac{t}{4}\bm{u}^\top \Lambda^{-1}[I - (I - \frac{1}{2t}\Lambda^{-1})^{-1}] \bm{u} \\&+ t[c-\frac{1}{4}\bm{u}^\top \Lambda^{-1} \bm{u}] \} \exp\{-t[\gamma + \Tr(\Lambda) + \|b\|_2 + c]\} \\ &|I-2t\Lambda|^{\frac{-1}{2}}\\
    &= \frac{\exp \{\frac{-t}{4}\bm{u}^\top \Lambda^{-1}(I - \frac{1}{2t}\Lambda^{-1})^{-1} \bm{u} + tc\}}{\exp\{t[\gamma + \Tr(\Lambda) + \|b\|_2 + c]\} |I-2t\Lambda|^{\frac{1}{2}}}
    \leq \\ 
    &\exp \{\frac{-t}{4\|\bm{b}\|_2^2} \lambda_{\min}(\Lambda^{-1}(I - \frac{1}{2t}\Lambda^{-1})^{-1}) \\&- t(\gamma + \Tr(\Lambda) + \|b\|_2 )\} |I-2t\Lambda|^{\frac{-1}{2}}\\
    &= \exp \{\frac{-t}{4\|\bm{b}\|_2^2} \frac{1}{\|\Lambda\| - 1/(2t)} - t(\gamma + \Tr(\Lambda) + \|b\|_2 )\} \\ &|I-2t\Lambda|^{\frac{-1}{2}}
    \leq \frac{\exp \{\frac{-t}{4\|\bm{b}\|_2^2 \|\Lambda\|} - t(\gamma + \Tr(\Lambda) + \|b\|_2 )\}}{|I-2t\Lambda|^{\frac{1}{2}}}.
\end{align*} \hfill $\square$

\subsection{Proof for Theorem \ref{theorem:funbound}}
\label{proof:funbound}

Note that if $\tilde{P}(\bm{x}) < 0$, the classifier predicts a label for class -1, else the predicted label would be 1. Here, $\bm{x} = y\bm{\mu} + \bm{z}$ where $\bm{z} \sim \mathcal{N}(\bm{0}, I)$ and $y\in\{\pm1\}$ since $(\bm{x}, y) \sim \mathcal{D}$.

\begin{align*}
    &\tau_{\mathcal{D}}(\tilde{P}) =~ \mathbb{E}\{\mathbbm{1}(y(\bm{x}^{\top} A\bm{x} + \bm{b}^{\top} \bm{x} + c) > 0)\}\\
    &=~ \mathbb{P}\{y(\bm{\mu}^{\top} A \bm{\mu} + \bm{z}^{\top} A \bm{z} + 2y\bm{\mu}^{\top} A \bm{z} + 
 \\ &y\bm{b}^{\top} \bm{\mu} + \bm{b}^{\top} \bm{z} + c) > 0\} \\
    &=~ \mathbb{P}(y=1) ~\mathbb{P}\{y(\bm{\mu}^{\top} A \bm{\mu} + \bm{z}^{\top} A \bm{z} \\&+ 2y\bm{\mu}^{\top} A \bm{z} + y\bm{b}^{\top} \bm{\mu} + \bm{b}^{\top} \bm{z} + c) > 0 ~|~ y=1\} ~+\\
     &\mathbb{P}(y=-1) ~\mathbb{P}\{y(\bm{\mu}^{\top} A \bm{\mu} + \bm{z}^{\top} A \bm{z} \\&+ 2y\bm{\mu}^{\top} A \bm{z} + y\bm{b}^{\top} \bm{\mu} + \bm{b}^{\top} \bm{z} + c) > 0 ~|~ y=-1\} =\\
    &{\frac{1}{2} \mathbb{P}\{\bm{z}^{\top} A \bm{z} + (\bm{b} + 2A\bm{\mu})^{\top} \bm{z} + \bm{\mu}^{\top} A \bm{\mu} + \bm{b}^{\top} \bm{\mu} + c > 0\}}~+\\
    &{\frac{1}{2} \mathbb{P}\{- \bm{z}^{\top} A \bm{z} - (\bm{b} - 2 A\bm{\mu})^{\top} \bm{z} -\bm{\mu}^{\top} A \bm{\mu} + \bm{b}^{\top} \bm{\mu} - c > 0\} } \\&:= p_1 + p_2
\end{align*}

We can see that
\begin{align*}
    &-\gamma_1 :=\\ &\mathbb{E}\{\bm{z}^{\top} A \bm{z} + (\bm{b} + 2A\bm{\mu})^{\top} \bm{z} + \bm{\mu}^{\top} A \bm{\mu} + \bm{b}^{\top} \bm{\mu} + c\} \\ &=~\Tr(\Lambda) + \|\bm{b} + 2A\bm{\mu}\|_2 + \bm{\mu}^\top A \bm{\mu} + \bm{b}^\top \bm{\mu} + c, ~~\textrm{and}\\
    &-\gamma_2 :=\\ &\mathbb{E}\{-\bm{z}^{\top} A \bm{z} - (\bm{b} - 2A\bm{\mu})^{\top} \bm{z} - \bm{\mu}^{\top} A \bm{\mu} + \bm{b}^{\top} \bm{\mu} - c\} \\ &=~ -\Tr(\Lambda) + \|\bm{b} - 2A\bm{\mu}\|_2 - \bm{\mu}^\top A \bm{\mu} + \bm{b}^\top \bm{\mu} - c.
\end{align*}

Using Lemma \ref{lemma:chernoff}, with $\gamma = \gamma_1, t=t_1$ for computing $p_1$ and $\gamma = \gamma_2, t=t_2$ for computing $p_2$ where $t_1, t_2$ are some non-negative constants,  we get

\begin{align*}
    &p_1 = \frac{1}{2|I - 2t_1 \Lambda|^{1/2}} \exp \bigg[ t_1 \bigg( \bm{\mu}^\top A \bm{\mu} + \bm{b}^\top \bm{\mu} + c \\&- \frac{1}{4 \|2A\bm{\mu} + \bm{b}\|_2 \|\Lambda\|} \bigg) \bigg], ~~\textrm{and}\\
    &p_2 = \frac{1}{2|I - 2t_2 \Lambda|^{1/2}} \exp \bigg[ t_2 \bigg( -\bm{\mu}^\top A \bm{\mu} + \bm{b}^\top \bm{\mu} - c \\&- \frac{1}{4 \|2A\bm{\mu} - \bm{b}\|_2 \|\Lambda\|} \bigg) \bigg].
\end{align*}

This gives us the upper bound for $\tau_{\mathcal{D}}(\tilde{P})$. However, to make sure that this upper bound is smaller than 1, we need to assert more conditions. $p_1$ and $p_2$ become smaller as $\gamma_1$ and $\gamma_2$ are larger positive numbers. However, $\gamma_1 + \gamma_2 = -(\|2A\bm{\mu} + \bm{b}\|_2 + \|2A\bm{\mu} - \bm{b}\|_2 + 4 \bm{\mu}^\top Q^\top (D_1^{-1} + D_{-1}^{-1}) Q \bm{\mu} ) \leq 0$ since $(D_1^{-1} + D_{-1}^{-1}) \succcurlyeq 0$. Hence, we look at separately at cases when either $\gamma_1 > 0$ or $\gamma_2 > 0$. 

If $\gamma_1 > 0$, then $\tau_{\mathcal{D}}(\tilde{P}) = \frac{1}{2}(p_1 + 1) < 1$. Else, if $\gamma_2 > 0$, then $\tau_{\mathcal{D}}(\tilde{P}) = \frac{1}{2}(p_2 + 1) < 1$. We know that for $\bm{\mu} \neq \bm{0}$, $\tau_{\mathcal{D}}({P}) = \phi(\bm{\mu}) > \frac{1}{2}$. Moreover, for any $a_{-1} \in [0, 0.5]$, $\exists a_1$ such that $\tau_{\mathcal{D}}(\tilde{P}) < \tau_{\mathcal{D}}({P})$. This can be satisfied by picking $a_1$ such that either $\gamma_1$ or $\gamma_2$ is very large, i.e., $\frac{1}{2} < \tau_{\mathcal{D}}(\tilde{P})=  \frac{1}{2}[1 + \min (p_1, p_2)] < \tau_{\mathcal{D}}({P})$. We note that the conditions $-\gamma_1 = \bm{\mu}^\top A \bm{\mu} + \bm{b}^\top \bm{\mu} + c + \Tr(A) + \|2A\bm{\mu} + \bm{b}\|_2 < 0$ and $-\gamma_2 = -\bm{\mu}^\top A \bm{\mu} + \bm{b}^\top \bm{\mu} - c - \Tr(A) + \|2A\bm{\mu} - \bm{b}\|_2 < 0$ can always be satisfied by picking a sufficiently large $\bm{\mu}$ in the direction of an eigenvector corresponding to a negative eigenvalue of $A$ (note that $A$ has negative eigenvalues). \hfill $\square$

\subsection{Details on generating Figure \ref{fig:theory}}
\label{app:analysis}

We use $\bm{\mu} \in \mathbb{R}^{d}, ~d=100$ to generate clean dataset with 1000 data points. They are randomly split into training and testing partitions of equal size. All the assumptions are consistent with the details provided  in the main body. We use $30 \times 30$ mesh-grid to plot the contour plots. While plotting the theoretical upper bounds, we choose the best $t_1, t_2$ with grid search from a search space $[2^1, 2^{0}, 2^{-1}, 2^{-2}, 2^{-3}, 2^{-4}, 2^{-5}]$.

\subsection{Experimental details}
\label{app:details}

This subsection provides the details for experiments in Section \ref{sec:experiments}.

\textbf{Hardware.} We use NVIDIA\textsuperscript{\textregistered} RTX A4000 GPU with 16GB memory with 16 AMD\textsuperscript{\textregistered} EPYC 7302P CPU cores. 

\textbf{Data augmentations.} For CIFAR-10 and CIFAR-100, we use random flipping, 4 pixel padding, and random $32\times 32$ size cropping. For ImageNet-100, we use random flipping and random cropping with resizing to $224\times 224$ size. All the images are rescaled to have pixel values in the range $[0, 1]$.

\textbf{Baselines.} We compare CUDA against error-minimizing noise \cite{em}, targeted adversarial poisoning \cite{tap}, neural tangent generalization attack \cite{ntga}, and robust error-minimizing noise \cite{rem}. We use the experimental outputs reported in \cite{rem} for our comparisons. For REM we choose $\rho_u = 8/255$ and $\rho_a = 4/255$ since REM works the best when $\rho_u = 2 \rho_a$ \cite{rem}. We perform experiments on REM not present in their work using their code available publicly on GitHub \footnote{\url{https://github.com/fshp971/robust-unlearnable-examples}} (MIT License).

\textbf{Networks.} For consistency, we use the same architectures used in \cite{rem}. We use their GitHub script\footnote{\url{https://github.com/fshp971/robust-unlearnable-examples/tree/main/models}} for this purpose.

\textbf{Training.} We train all the networks for 100 epochs. The initial learning rate is 0.1. Learning rate decays to 0.01 at epoch 40 and to 0.001 at epoch 80. We use a stochastic gradient descent optimizer with a momentum factor of 0.9, weight decay factor of 0.0005, and batch size of 128. For adversarial training, we follow the procedure in \cite{adversarialtraining}. We use 10 steps of projected gradient descent with a step size of $0.15 \rho_a$. 

\textbf{Analysis of CUDA.} For grayscaling experiments, we use images with their average channel values as the input to the network. Test accuracy is computed on the grayscaled test datasets. For smoothing, we use the GitHub codes\footnote{\url{https://github.com/Hadisalman/smoothing-adversarial}} from \cite{smoothingcohen} (MIT License). For mixup \cite{mixup}, we use the default value of $\alpha = 1.0$.

\begin{figure*}[t]
 \centering
     \includegraphics[width=0.8\textwidth]{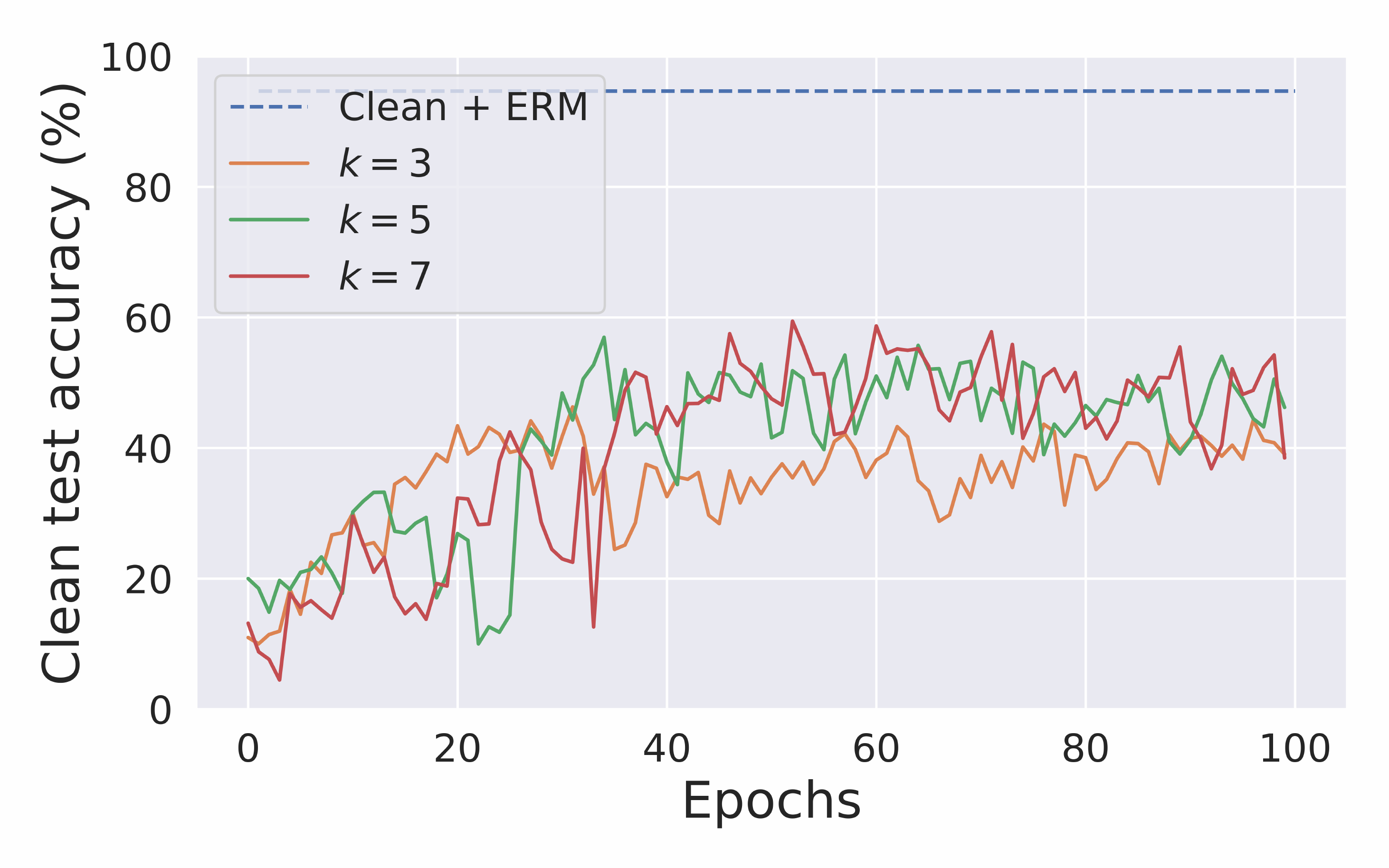}
     \caption{CUDA CIFAR-10 images ($k=3, p_b=0.3$) trained using ResNet-18 with the Deconvolution-based Adversarial Training framework with varying transpose convolution filter sizes $k$.}
     \label{fig:fat}
 \end{figure*}

\textbf{Deconvolution-based adversarial training (DAT).} We experiment with various filter sizes of 3,5, and 7 for the transpose convolution filters. For each batch of data, we use 10 steps of projected gradient descent with a learning rate of 0.1 to learn  transpose convolution filters for each class. The weights and biases of the transpose convolution filters are constrained to be within $[-C, C]$. We choose $C=5$. After 10 steps of inner maximizing optimization, the resulting image is rescaled such that the pixel values lie in $[0, 1]$. See Figure \ref{fig:fat} for clean test accuracy vs. epochs plot for DAT with varying transpose filter sizes. As seen in Figure \ref{fig:funcifar}, DAT can break CUDA CIFAR-10 with a low blur parameter value of $p_b=0.1$ to get a clean test accuracy $\sim$78$\%$. However, with higher $p_b$ values DAT can not achieve more than 50$\%$ clean test accuracy. DAT solves the following optimization problem:

\begin{align}
\label{opt:fat}
    \arg \min_{\theta} \frac{1}{n} \sum_{k \in [K]} \max_{\|s_{k}\|_{\infty} \leq C} \sum_{i : y_i = k} \ell (f_\theta(\mathbf{x}_i \star s_{k}), k)
\end{align}

where $\star$ denotes the transpose convolution operator, $s_{y_i}$ denotes the transpose convolution filter for class $y_i$, and $\ell$  is the soft-max cross-entropy loss function.

  \begin{figure*}[t]
 \centering
     \includegraphics[width=0.8\textwidth]{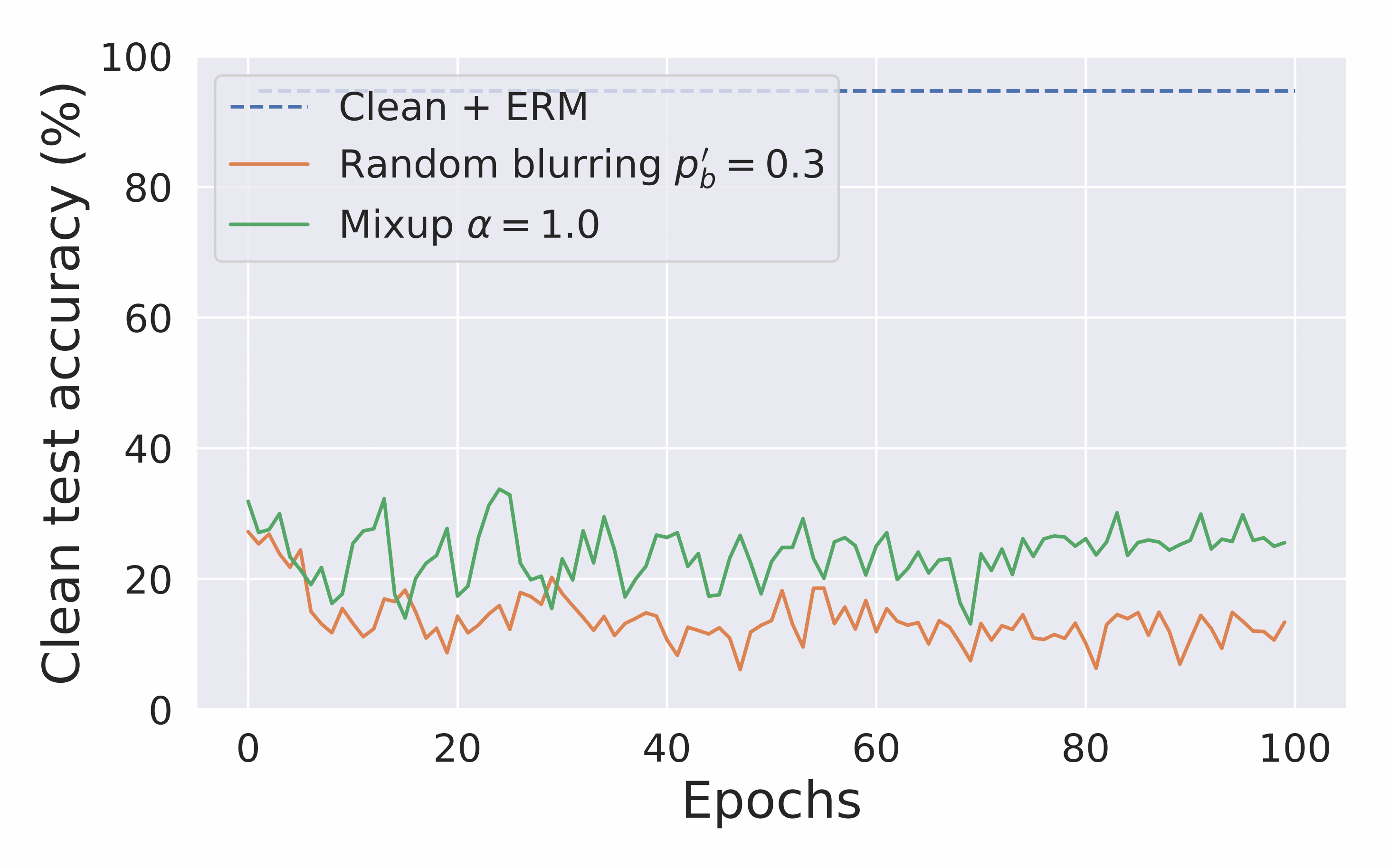}
     \caption{CUDA CIFAR-10 images ($k=3, p_b=0.3$) trained using ResNet-18 with mixup and random blurring augmentations.}
     \label{fig:augs}
 \end{figure*}

\textbf{CUDA with augmentations.} We use mixup with the default $\alpha = 1.0$ \cite{mixup}. See Figure \ref{fig:augs} for the training curve. For random blurring augmentations, we use $p_b' = 0.1, 0.3$ and $k=3$. With both these parameters, CUDA is seen to be effective. See Figure \ref{fig:augs} for the training curve with $p_b' = 0.3$.

 \subsection{More experimental results}
\label{app:more}

\begin{figure*}
     \centering
     \begin{subfigure}[b]{0.9\textwidth}
         \centering
         \includegraphics[width=0.9\textwidth]{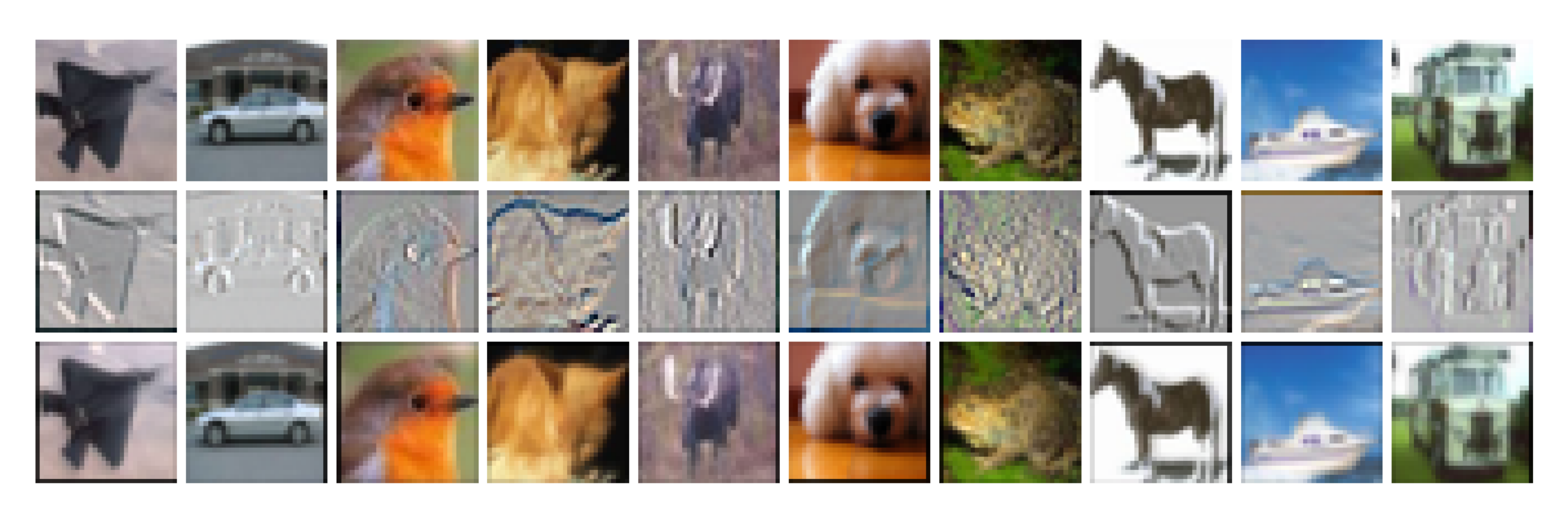}
         \caption{$p_b = 0.1$}
     \end{subfigure}
     \\
     \begin{subfigure}[b]{0.9\textwidth}
         \centering
         \includegraphics[width=0.9\textwidth]{suppl_images_pdfs/sample_0.1.pdf}
         \caption{$p_b = 0.3$}
     \end{subfigure}
     \\
     \begin{subfigure}[b]{0.9\textwidth}
         \centering
         \includegraphics[width=0.9\textwidth]{suppl_images_pdfs/sample_0.1.pdf}
         \caption{$p_b = 0.5$}
     \end{subfigure}
        \caption{CUDA CIFAR-10 images generated using different blur parameters $p_b$. The top row shows the clean images, the bottom row shows the corresponding CUDA image, and the middle row shows the normalized difference between the clean and the CUDA image.}
        \label{fig:fun_pbs_figs}
\end{figure*}

\begin{figure*}
     \centering
     \begin{subfigure}[b]{0.9\textwidth}
         \centering
         \includegraphics[width=0.9\textwidth]{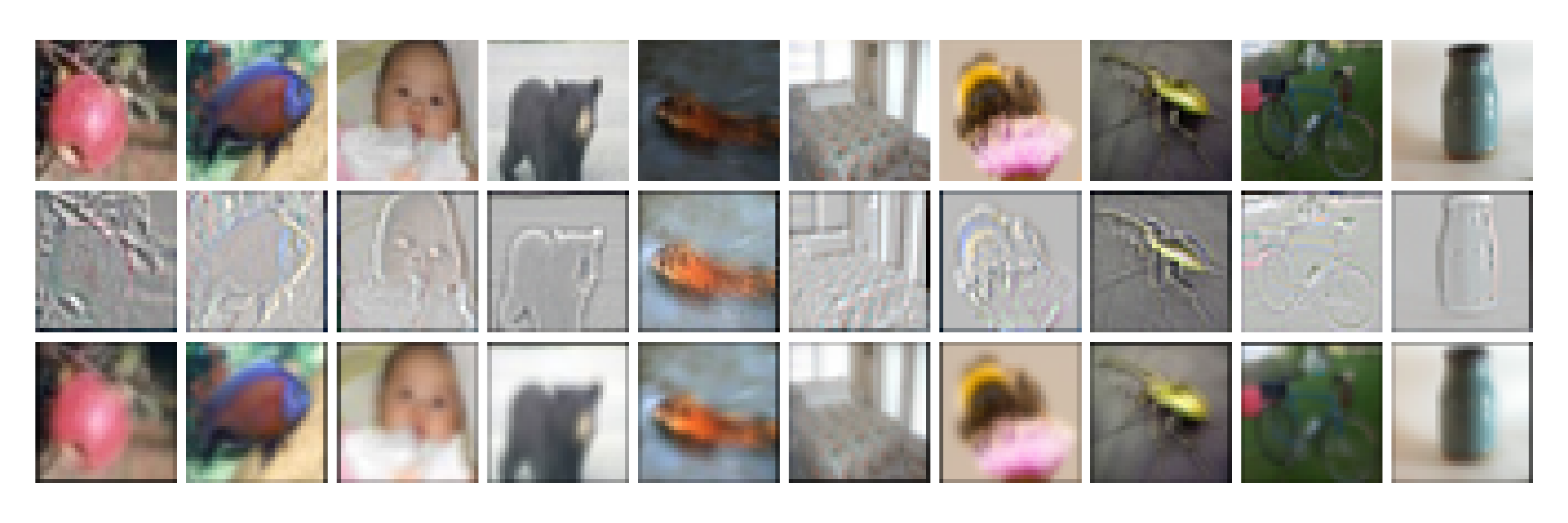}
         \caption{CIFAR-100}
     \end{subfigure}
     \\
     \begin{subfigure}[b]{0.9\textwidth}
         \centering
         \includegraphics[width=0.9\textwidth]{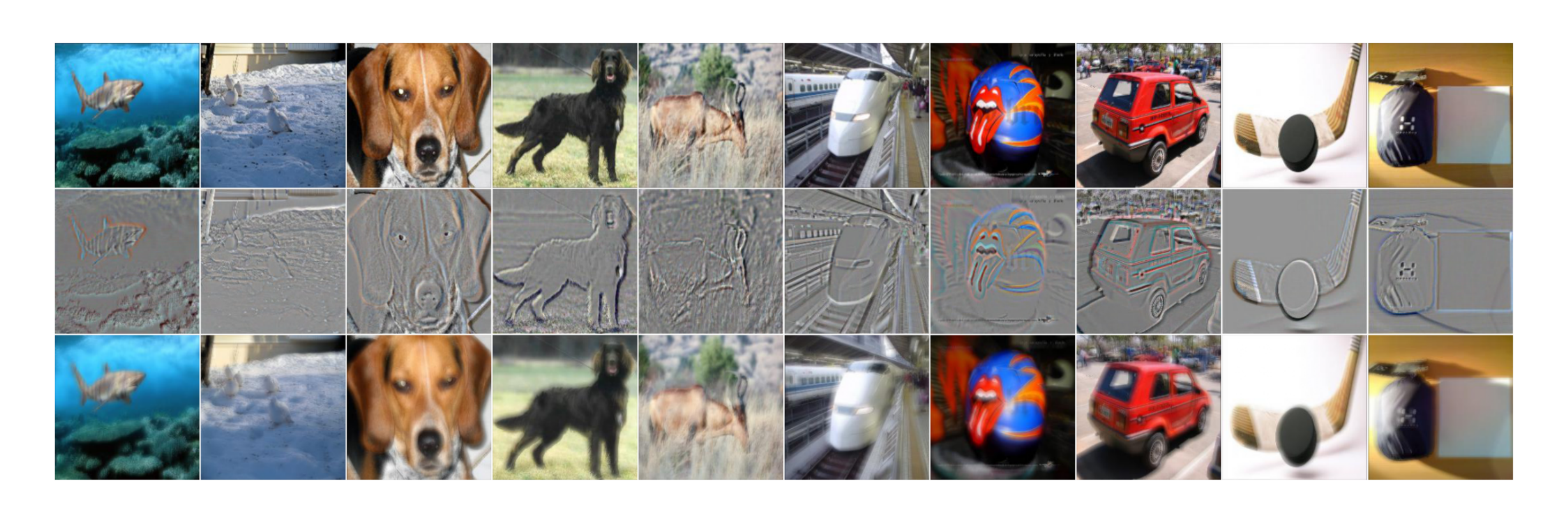}
         \caption{ImageNet-100}
     \end{subfigure}
        \caption{CUDA CIFAR-100 and ImageNet-100 images generated using $k=3, p_b = 0.3$ and $k=9, p_b = 0.06$, respectively. The top row shows the clean images, the bottom row shows the corresponding CUDA image, and the middle row shows the normalized difference between the clean and the CUDA image.}
        \label{fig:fun_pbs_figs_1}
\end{figure*}

\begin{figure*}[t]
 \centering
     \includegraphics[width=0.8\textwidth]{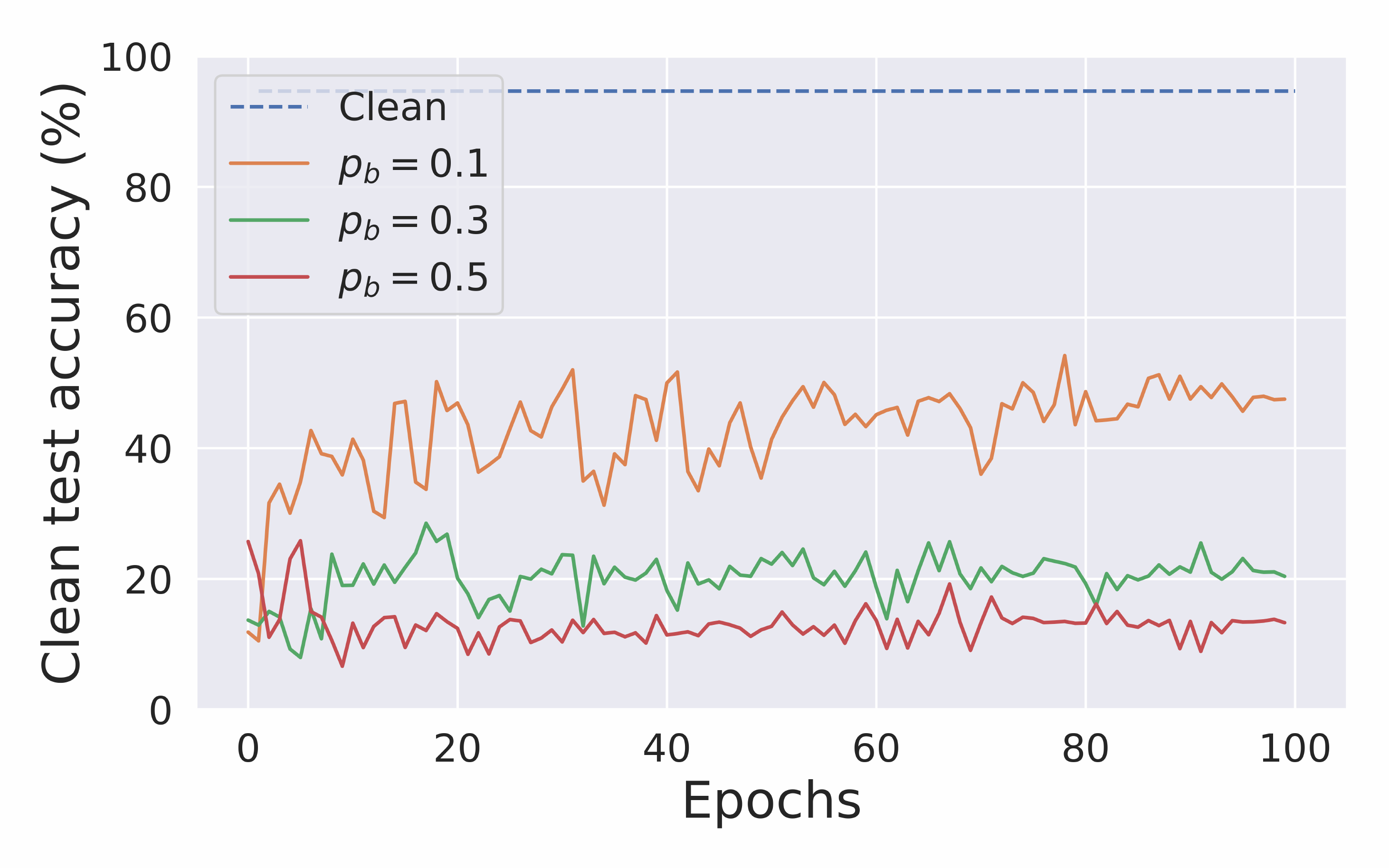}
     \caption{ResNet-18 trained using CUDA CIFAR-10 data generated using different blur parameters $p_b$.}
     \label{fig:fun_pbs}
 \end{figure*}

\begin{figure*}[t]
 \centering
     \includegraphics[width=0.8\textwidth]{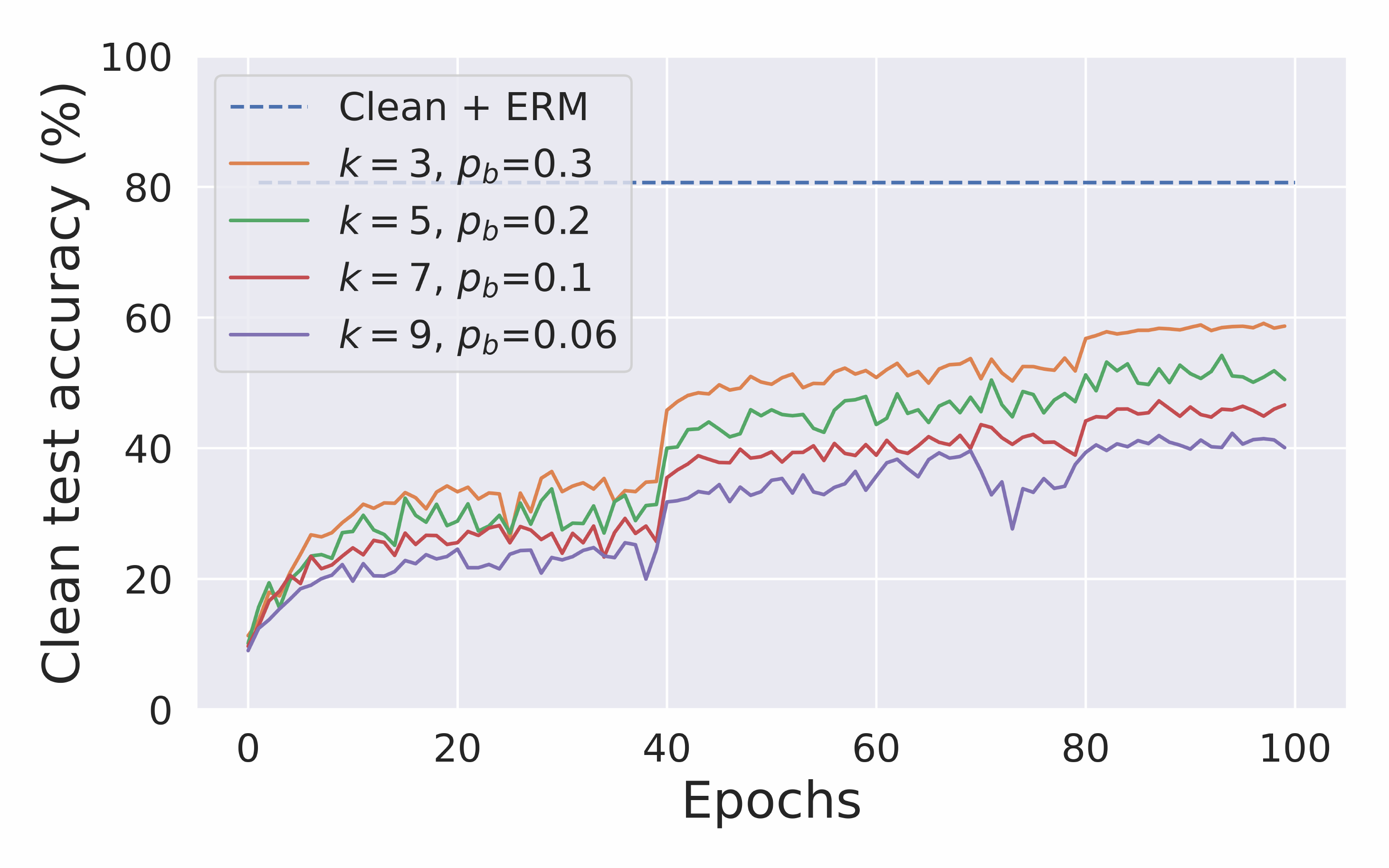}
     \caption{ResNet-18 trained using CUDA ImageNet-100 dataset generated using different filter sizes $k$.}
     \label{fig:fun_params}
 \end{figure*}

\begin{figure*}
     \centering
     \begin{subfigure}[b]{0.32\textwidth}
         \centering
         \includegraphics[width=1\textwidth]{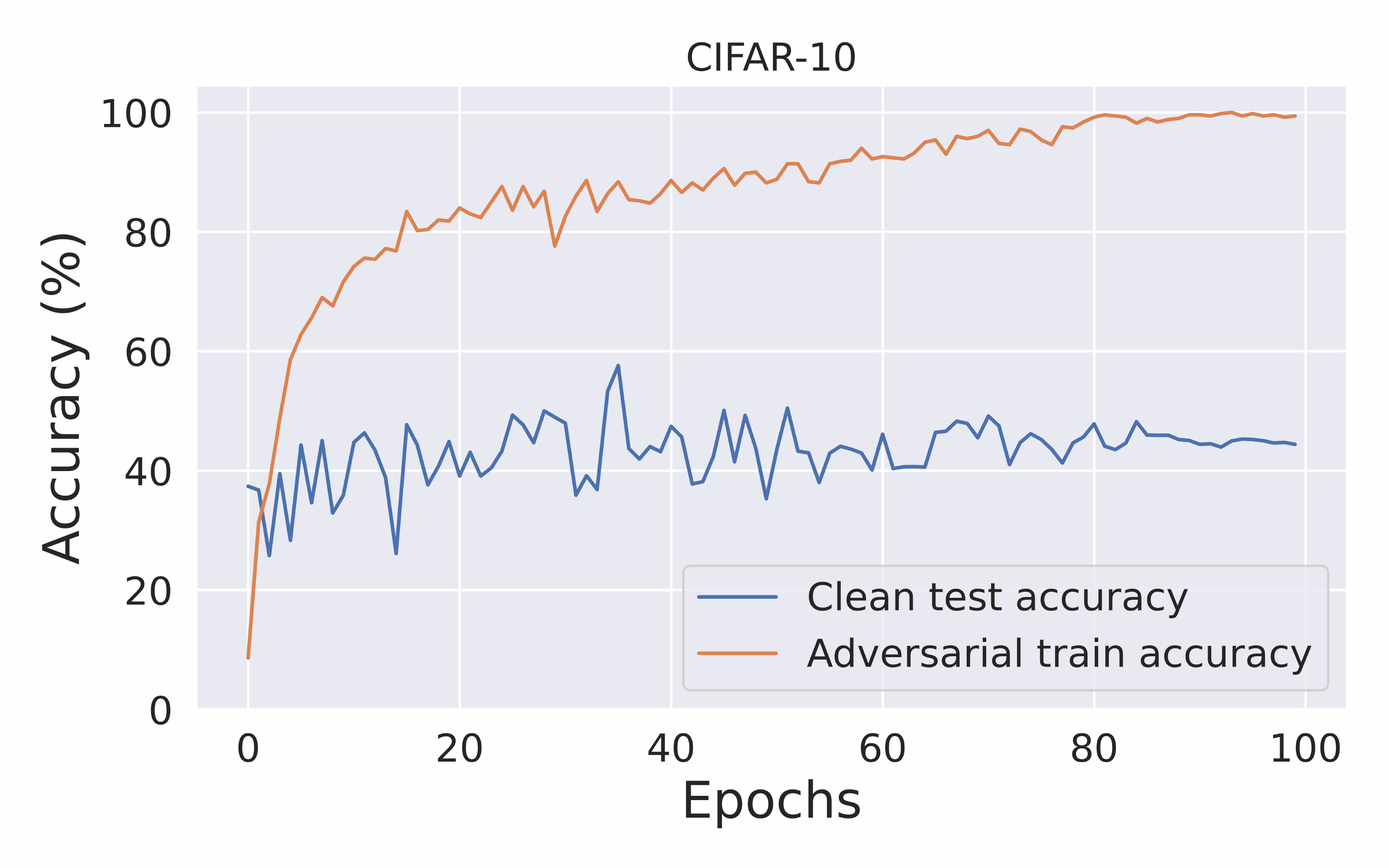}
         \caption{CIFAR-10}
     \end{subfigure}     \hfill
     \begin{subfigure}[b]{0.32\textwidth}
         \centering
         \includegraphics[width=1\textwidth]{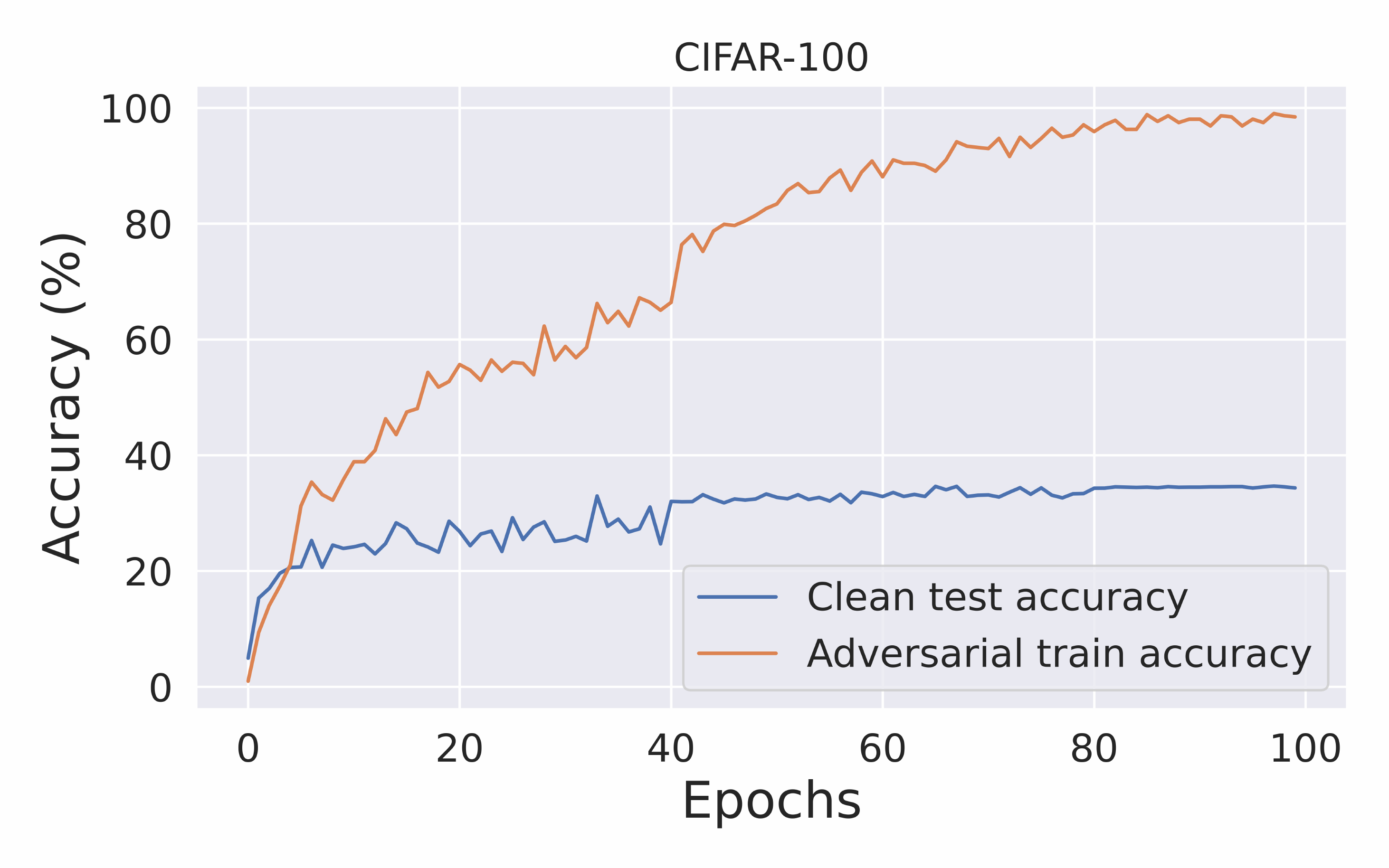}
         \caption{CIFAR-100}
     \end{subfigure}     \hfill
     \begin{subfigure}[b]{0.32\textwidth}
         \centering
         \includegraphics[width=1\textwidth]{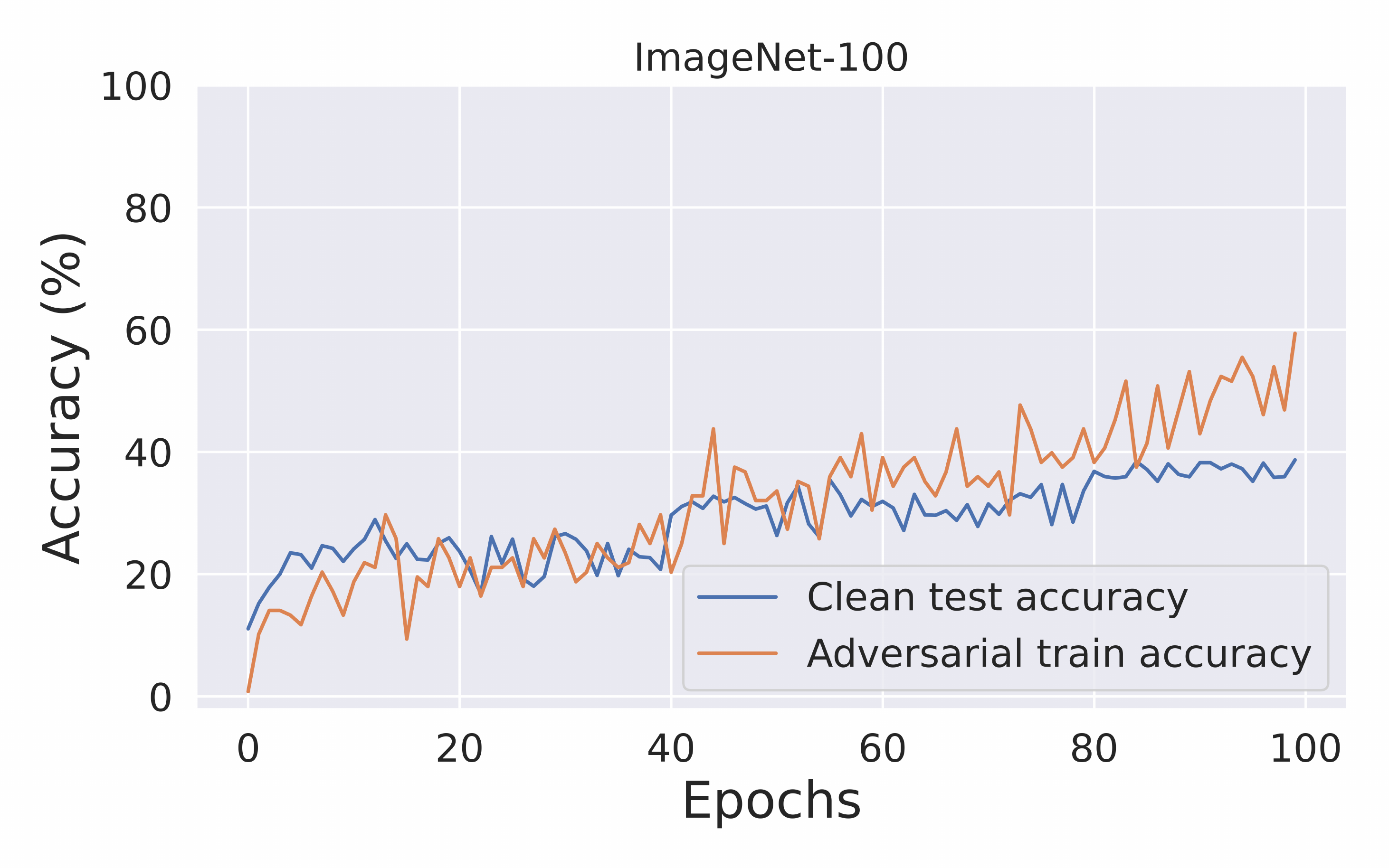}
         \caption{ImageNet-100}
     \end{subfigure}
        \caption{Adversarial training curves for ResNet-18 with CIFAR-10, CIFAR-100, and ImageNet-100 CUDA datasets.}
        \label{fig:fun_at_curves}
\end{figure*}

Figure \ref{fig:fun_pbs_figs} shows the CUDA CIFAR-10 data generated using $k=3$ and different $p_b$ blur parameters.  Figure \ref{fig:fun_pbs_figs_1} shows the CUDA CIFAR-100 and CUDA ImageNet-100 data generated using $k=3, p_b = 0.3$ and $k=9, p_b = 0.06$, respectively.  Figure \ref{fig:fun_pbs} shows the clean test accuracy of ResNet-18 with CUDA CIFAR-10 generated using different blur parameters. As we see in the plots, higher the blur parameter, better the effectiveness of CUDA is. However, we choose $p_b = 0.3$ for our experiments since the the images generated using this hyperparameter look perceptibly more similar to the clean images (when compared to $p_b = 0.5$) while giving a very low clean test accuracy. A lower value of $p_b=0.1$ gives better unlearnability. However, CUDA CIFAR-10 generated using $p_b=0.1$ is not robust with our Deconvolution-based Adversarial Training, as shown in Figure \ref{fig:funcifar}.  Figure \ref{fig:fun_params} shows the clean test accuracy of ResNet-18 with CUDA ImageNet-100 dataset generated using different filter sizes. Figure \ref{fig:fun_at_curves} shows the adversarial training curves for ResNet-18 with different CUDA datasets.

We show the effectiveness of CUDA with Tiny-ImageNet \cite{le2015tiny}, DeIT \cite{deit}, EfficientNetV2 \cite{en}, and MobileNetV2 \cite{mn} below.

\begin{table}[!h]
    \centering \footnotesize
    \begin{tabular}{|c|c|c|} \hline
         Model & ERM & $L_2$ AT ($\epsilon=0.5$)  \\ \hline
         DeIT \cite{deit} &  24.85 $\%$ & 38.90 $\%$ \\
         EfficientNetV2-S \cite{en} & 20.47 $\%$ & 42.19 $\%$ \\
         MobileNetV2 \cite{mn} & 21.10 $\%$ & 32.00 $\%$ \\ \hline
    \end{tabular}
    \caption{Effectiveness of CUDA on CIFAR-10.}
    \label{tab:cnnmodels}
\end{table}

\begin{table}[!h]
    \centering \footnotesize
    \begin{tabular}{|c|c|c|}
    \hline
         Training method & Clean & CUDA  \\
         \hline
         ERM & 48.14 $\%$ & 5.98 $\%$\\ 
         $L_2$ AT ($\epsilon=0.5$) & 42.72 $\%$ & 14.54 $\%$ \\
         \hline
    \end{tabular}
    \caption{Effectiveness of Tiny-ImageNet CUDA with ResNet-18. We use the same hyperparameters as our CIFAR experiments.}
    \label{tab:tinyimgnet}
\end{table}

 \subsection{Effects of blurring}
 \label{app:blur1}
 
\begin{figure*}
     \centering
     \begin{subfigure}[b]{0.45\textwidth}
         \centering
         \includegraphics[width=1\textwidth]{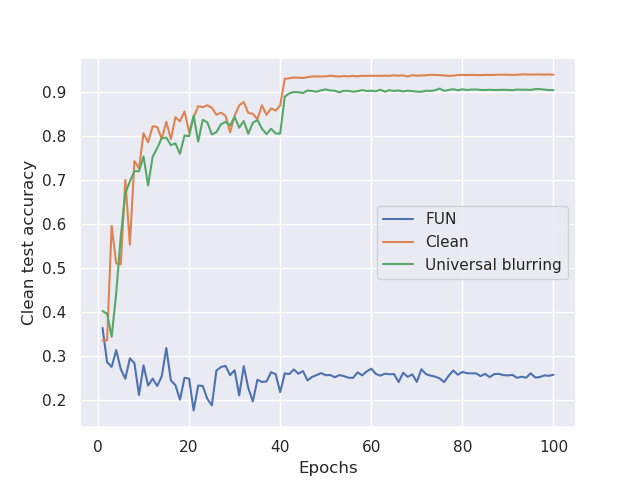}
         \label{fig:uni_acc}
         \caption{Test accuracy}
     \end{subfigure}     
     \hfill
     \begin{subfigure}[b]{0.45\textwidth}
         \centering
         \includegraphics[width=1\textwidth]{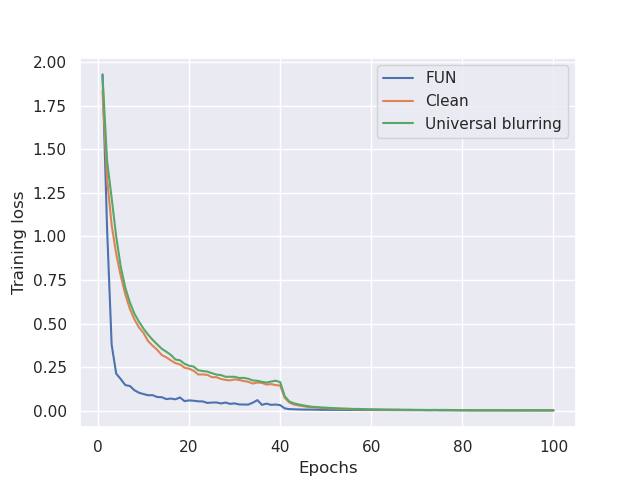}
         \label{fig:uni_loss}
         \caption{Training loss}
     \end{subfigure}    
        \caption{ResNet-18 trained using clean, CUDA, and universally blurred CIFAR-10 datasets.}
        \label{fig:blurring_effects}
\end{figure*}

 Here, we study the effects of blurring. We investigate if class-wise blurring is required for achieving unlearnability. For this, we use a universal filter (generated using the same $p_b=0.3$ and $k=3$ hyperparameters) to blur all the training images in the dataset. A ResNet-18 trained on this dataset achieves a clean test accuracy of 90.47$\%$. Essentially, the blurring that is performed only degrades the clean test accuracy by $\sim$4$\%$. This means that class-wise blurring (CUDA) is required for achieving the unlearnability effect (see Figure \ref{fig:blurring_effects}). This experiment also demonstrates that the blurring we perform does not make the dataset useless or destroy its semantics. For this experiment, we use models with fixed initialization and random seeds.

 \subsection{Why does CUDA work?}
 \label{app:blur2}
 
 In this section, we perform experiments that show that a model trained on CUDA dataset learns the relation between the class-wise filters and the labels. We train ResNet-18 using the CUDA CIFAR-10 dataset for the experiments. We perform three independent trials for each of the experiments and report the mean performance scores. Trained models achieve a mean clean test accuracy of 21.34$\%$. Now, we use the class-wise filters to perturb the images in the test set based on their corresponding labels. Trained models achieve a very high mean accuracy of 99.91$\%$ on this perturbed test set. This shows that the trained models learned the relation between the filters and their corresponding labels. Next, we permute the filters to perturb the test set such that test set images with label $i$ are perturbed with the filters of class $(i+1)\%10$. Trained models achieve a very low mean accuracy of 2.53$\%$ on this perturbed test set. This is evidence that CUDA can also be used for backdoor attacks.

\subsection{Effect of transfer learning}
\label{app:transfer}

\begin{figure*}
     \centering
     \begin{subfigure}[b]{0.45\textwidth}
         \centering
         \includegraphics[width=\textwidth]{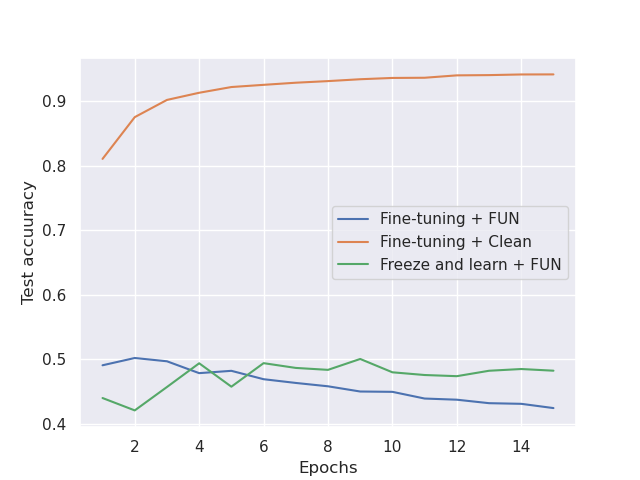}
         \label{fig:downstream_acc}
         \caption{Test accuracy}
     \end{subfigure}     
     \hfill
     \begin{subfigure}[b]{0.45\textwidth}
         \centering
         \includegraphics[width=\textwidth]{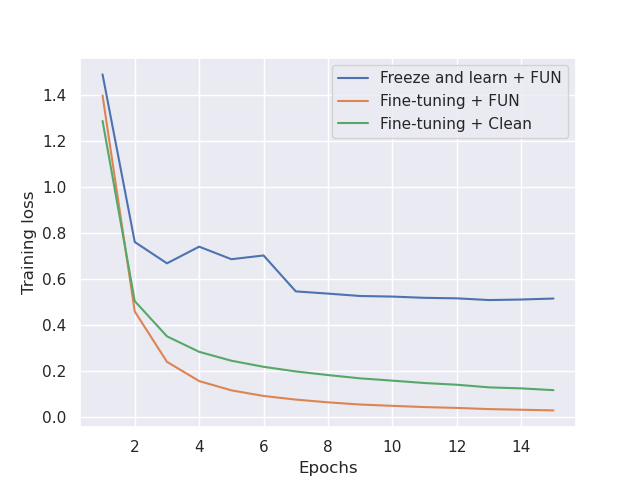}
         \label{fig:downstream_loss}
         \caption{Training loss}
     \end{subfigure}    
        \caption{Pre-trained ResNet-18 with fine-tuning and training the linear layer using CUDA and clean CIFAR-10 datasets.}
        \label{fig:downstream}
\end{figure*}

In this section, we experiment the effect of using a pre-trained ResNet-18 with PyTorch \cite{pytorch}. We train it on the CUDA CIFAR-10 dataset in two different ways. First, we fine-tune the whole network on the CUDA dataset with a learning rate of 0.001 for 15 epochs. This achieves a clean test accuracy of 42.42$\%$. Fine-tuning the network with clean training data gives 94.19$\%$ clean test accuracy. Next, we freeze all the layers except the final layer to train a linear classifier with the pre-trained weights using the CUDA CIFAR-10 dataset. We call this ``Freeze and learn". We use a SGD optimizer to train the linear layer for 15 epochs with an initial learning rate of 0.1. The learning rate is decayed by a factor of 10 after every 5 epochs. This achieves a clean test accuracy of 48.22$\%$. The results are shown in Figure \ref{fig:downstream}. This experiment shows that pre-trained network with CUDA data training does not help achieve good generalization on the clean data distribution. 

 \subsection{Effect of CUDA with regularization techniques}
 \label{app:regularize}
 
 In this section, we study the effect of training a ResNet-18 with CUDA CIFAR-10 dataset using various regularization techniques such as mixup \cite{mixup}, cutout \cite{cutout}, cutmix \cite{cutmix}, autoaugment \cite{autoaugment}, and orthogonal regularization \cite{orth}. We perform mixup, cutout, cutmix, autoaugment, and orthogonal regularization to achieve 25.53$\%$, 25.80$\%$, 26.93$\%$, 34.09$\%$, and 50.72$\%$. Even though these regularizations help in improving the vanilla ERM training, these networks still do not achieve good generalization on the clean data distribution. We use cutout using GitHub codes\footnote{\url{https://github.com/uoguelph-mlrg/Cutout/blob/master/util/cutout.py}} with \texttt{length=16} and \texttt{n\_holes=1}, cutmix using GitHub codes\footnote{\url{https://github.com/hysts/pytorch_cutmix/blob/master/cutmix.py}} with $\alpha=1$, autoaugment using PyTorch \cite{pytorch}, mixup using GitHub codes\footnote{\url{https://github.com/facebookresearch/mixup-cifar10/blob/main/train.py}} with $\alpha=1$, and orthogonal regularization using GitHub codes\footnote{\url{https://github.com/kevinzakka/pytorch-goodies}} with  \texttt{reg=1e-6} (all MIT licenses).

\subsection{Network parameter distribution}
\label{app:paramdistbn}

\begin{figure*}[t]
 \centering
\includegraphics[width=0.7\textwidth]{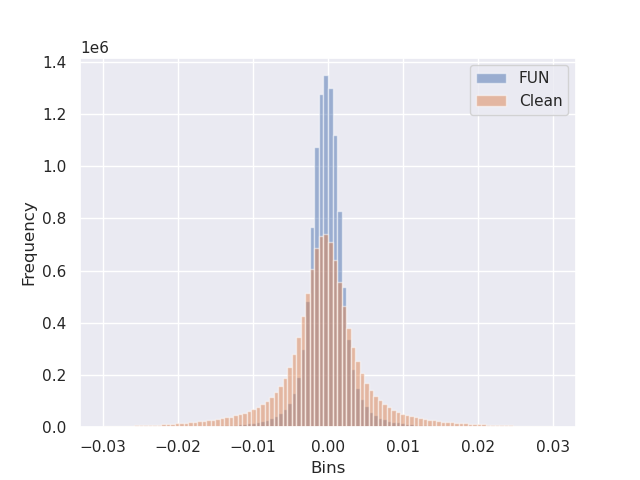}
     
     \caption{Network parameter distributions of ResNet-18 trained on clean and CUDA CIFAR-10 datasets.}
     \label{fig:dists}
 \end{figure*}

In this section, we compare the network parameter distributions of ResNet-18 trained on clean and CUDA CIFAR-10 datasets (see Figure \ref{fig:dists}). Both the distributions are similar to normal distributions with a mean of 0. However, the parameter distribution of the clean model has a higher standard deviation than the CUDA-based model’s parameter distribution.

\end{document}